\documentclass{article}

\usepackage[preprint]{neurips_2024}

\usepackage[utf8]{inputenc} 
\usepackage[T1]{fontenc}    
\usepackage{hyperref}       
\usepackage{url}            
\usepackage{booktabs}       
\usepackage{amsfonts}       
\usepackage{nicefrac}       
\usepackage{microtype}      
\usepackage{xcolor}         
\usepackage{multirow}
\usepackage{graphicx}
\usepackage{subfigure}

\title{WaterMamba: Visual State Space Model for Underwater Image Enhancement}

\author{%
   Meisheng Guan \\
   Ningbo University\\
   \texttt{meishengguan@gmail.com}
   \And
  Haiyong Xu\thanks{Corresponding author} \\
   Ningbo University\\
  \texttt{xuhaiyong@nbu.edu.cn}
     \And
       Gangyi Jiang \\
   Ningbo University\\
   \texttt{jianggangyi@nbu.edu.cn} \\
        \And
   Mei Yu \\
   Ningbo University\\
   \texttt{yumei@nbu.edu.cn} \\
        \And
   Yeyao Chen \\
   Ningbo University\\
   \texttt{chenyeyao@nbu.edu.cn} \\
        \And
   Ting Luo \\
   Ningbo University\\
   \texttt{luoting@nbu.edu.cn} \\
        \And
   Yang Song \\
   Ningbo University\\
   \texttt{songyang@nbu.edu.cn} \\
   }

\begin{document}

\maketitle

\begin{abstract}
  Underwater imaging often suffers from low quality and lack of fine details due to various physical factors affecting light propagation, scattering, and absorption in water. To improve the quality of underwater images, some underwater image enhancement (UIE) methods based on convolutional neural networks (CNN) and Transformer have been proposed. However, CNN-based UIE methods are limited in modeling long-range dependencies and are often applied to specific underwater environments and scenes with poor generalizability. Transformer-based UIE methods excel at long-range modeling, which typically involve a large number of parameters and complex self-attention mechanisms, posing challenges for efficiency due to the quadratic computational complexity to image size. Considering computational complexity and the severe degradation of underwater images, the state space model (SSM) with linear computational complexity for UIE, named WaterMamba, is proposed. Considering the challenges of non-uniform degradation and color channel loss in underwater image processing, we propose a spatial-channel omnidirectional selective scan (SCOSS) blocks consisting of the spatial-channel coordinate omnidirectional selective scan (SCCOSS) modules module and a multi-scale feedforward network (MSFFN). The SCOSS block effectively models pixel information flow in four directions and channel information flow in four directions, addressing the issues of pixel and channel dependencies. The MSFFN facilitates information flow adjustment and promotes synchronized operations within spatial-channel coordinate omnidirectional selective scan (SCCOSS) modules. Extensive experiments on four datasets showcase the cutting-edge performance of the WaterMamba while employing reduced parameters and computational resources. The WaterMamba outperforms the state-of-the-art method, achieving a PSNR of 24.7dB and an SSIM of 0.93 on the UIEB dataset. On the UCIOD dataset, the WaterMamba achieves a PSNR of 21.9dB and an SSIM of 0.84. Additionally, the UCIQE of the WaterMamba on the SQUID dataset is 2.77, while the UIQM is 0.56, which further validated its effectiveness and generalizability. The code will be released on GitHub after acceptance
.

\end{abstract}

\section{Introduction}
\begin{figure}
  \centering
\centerline{\includegraphics[height=4.3cm,width=6.9cm]{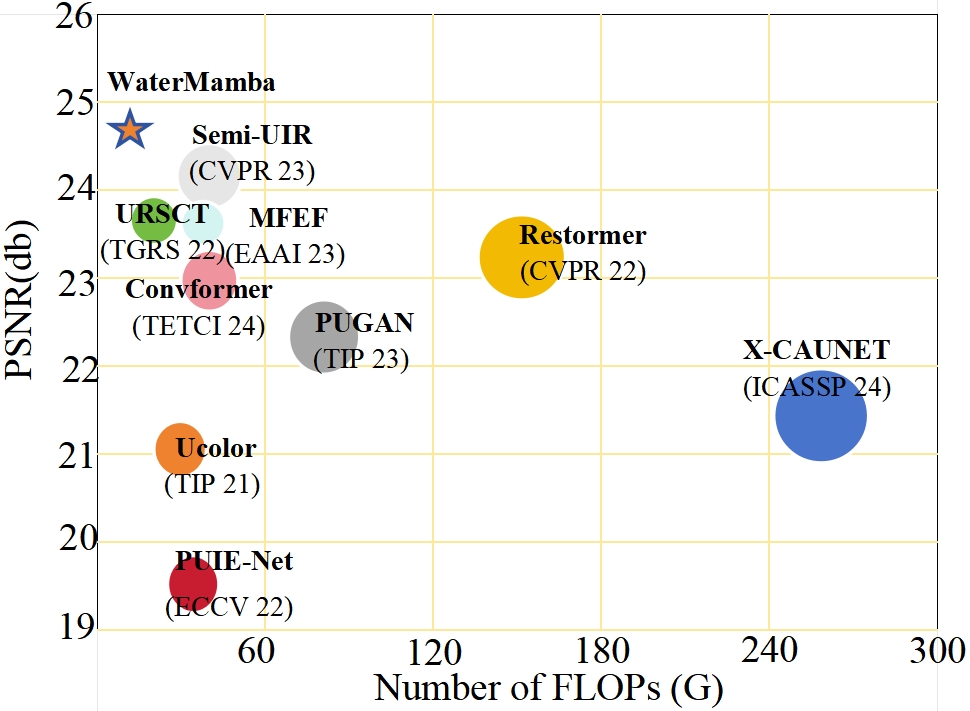}}\vspace{0.01cm}
  \caption{Quantitative comparison of FLOPs and parameter counts of WaterMamba with other state of the art underwater image enhancement and dehazing methods.}
  \label{fig1}
\end{figure}

	Underwater imaging plays a crucial role in enabling key capabilities for autonomous underwater vehicles (AUVs) \cite{1} and downstream vision-related tasks such as underwater target tracking \cite{2}, underwater robotics \cite{3}, and underwater biology exploration \cite{4}. However, the quality of underwater images is often degraded due to the optical properties of water that cause light absorption and scattering \cite{5}, which results in poor image contrast and visibility, posing challenges for vision-based detection and recognition. Enhancing the distinguishability and contrast of targets in underwater imagery would significantly improve the utility of visual data for downstream applications. Therefore, designing effective underwater image enhancement (UIE) methods is of paramount importance to unlock the full potential of underwater imaging systems and their diverse applications.

Recently, UIE methods aim to improve the quality and visibility of underwater images by recovering the absorbed and scattered color information have become an active research area. Generally, UIE methods can be categorized as traditional physical model-based and deep learning-based methods. Traditional physical model-based UIE methods \cite{6,7,8,9} have been proposed to address problems in underwater imaging such as color distortion and low contrast caused by light attenuation and scattering. However, these methods typically require complex parameter tuning and have limited robustness in practical applications. To address the issue, deep learning-based UIE \cite{10,11,12,13,14,15} methods have been proposed. In particular, convolutional neural networks (CNN) have been widely applied for UIE, which can automatically learn feature representations of images and perform end-to-end image restoration. However, CNN-based UIE methods have a limited size of the receptive field, making them unable to model long-range pixel dependencies effectively. Additionally, using static convolutional filters during inference cannot adapt to the variability in underwater images. 

To address these issues, the self-attention mechanism-based Transformer \cite{16}, which is first proposed for natural language processing (NLP) \cite{16} and later extended to various vision tasks with great success \cite{17,18}. However, the Transformer exhibits quadratic complexity with respect to sequence length, posing challenges for end-to-end  UIE requiring the processing of high-resolution images.

More recently, structured state space models (SSM) \cite{19}, such as Mamba, have been proposed to address the quadratic complexity issue of the TransformerHowever, existing visual Mamba for computer vision, while enabling full data modeling and multi-scale information, typically apply global max/average pooling to obtain channel attention, losing spatial information important for underwater images. Underwater images are prone to color distortion and loss of details due to light absorption and scattering by water molecules, especially in longer wavelengths like red and infrared. Additionally, light weakens and colors distort, resulting in bluish-green tinted images with loss of details in deep water. To address color distortion and pixel detail blurring in underwater images, we propose a novel WaterMamba method based on SSM with linear complexity. It introduces a spatial-channel omnidirectional selective scan (SCOSS) block containing a spatial-channel coordinate omnidirectional selective scan (SCCOSS) module and multi-scale feedforward network  (MSFFN). The SCOSS block embeds spatial attention into channel attention, while MSFFN effectively exploits multi-scale features to model local pixel relations. By stacking SCOSS blocks into U-Net, WaterMamba achieves state-of-the-art (SOTA) (see Fig. ~\ref{fig1} and Table ~\ref{tab1}) performance on multiple underwater datasets with lower computation and parameters. The main contributions of this work can be summarized as follows:
\begin{itemize}
\item[$\bullet$]Considering the quadratic computational complexity issues of Transformer, a SSM-based UIE method with linear complexity called WaterMamba is proposed. SCOSS blocks are stacked within a U-Net architecture by WaterMamba to reconstruct image details and colors. The enhanced underwater image is obtained via a residual connection at last.
\item[$\bullet$]Given that underwater images suffer from non-uniform degradation and severe color channel loss, the SCOSS block consisting of the SCCOSS module and an MSFFN module is designed. The multi-scale receptive fields of the MSFFN module can also model the non-uniform degradation in underwater images, improving the robustness of the model.
\item[$\bullet$]Extensive empirical evaluation was conducted on four benchmark underwater image datasets. The results demonstrate that the proposed WaterMamba algorithm achieves superior performance under lower computational and parameter requirements compared to existing methods. 
\end{itemize}

\section{Related Work}
\subsection{CNN-based Underwater Image Enhancement}
Recently, CNN-based UIE \cite{15,25,26,27} has become a major research topic. Compared to traditional physically model-based methods \cite{6,7,8}, CNN-based UIE methods can automatically learn effective feature representations from underwater images through an end-to-end learning process without relying on hand-crafted features. Li et al. \cite{10} proposed an underwater scene prior guided CNN method for enhancement without estimating underwater imaging model parameters but directly restoring clear diver images. Furthermore, Li et al. \cite{13} introduced a UIE network (Ucolor) with multi-color space embedding guided by medium transmission, designing an encoder network guided by medium transmission to enhance the network's response to degraded regions. Fu et al. \cite{14} proposed a novel probabilistic network to learn the enhancement distribution of degraded underwater images by combining a conditional variational autoencoder with adaptive instance normalization. Additionally, Cong et al.\cite{25} introduced a GAN method guided by physical models for UIE, designing subnetworks to learn the parameters for inverting physical models and introducing them to the discriminator to improve image realism and visual aesthetics. However, the local receptive field mechanism of traditional CNN makes it difficult to capture long-range pixel correlations in underwater images, thus limiting their capability for global information modeling. Additionally, CNN-based UIE methods often involve a large number of parameters and complex network structures, presenting challenges for real-time applications due to high computational overhead.

\subsection{Transformer-based Underwater Image Enhancement}
To address the issue, Transformer-based UIE \cite{28,30,33} methods have also attracted significant attention in recent years. Compared to CNN, the self-attention mechanism in Transformer can capture long-range pixel correlations in input images, overcoming the limitation of local receptive fields in CNN to achieve effective global information modeling. This is crucial for modeling complex scenes of the underwater image \cite{5}.

Recently, Ren et al. \cite{29} proposed a Swin Transformer method for UIE that fuses convolution and attention mechanisms by strengthening local attention to enhance modeling capacity. Huang et al. \cite{30} designed an adaptive group attention mechanism that can dynamically select vision-complementary channels based on dependencies, reducing the number of attention parameters. Shen et al. \cite{31} proposed a UIE method named UDAformer based on dual attention transformations, combining channel self-attention and pixel self-attention mechanisms. Peng et al. \cite{32} designed a U-shaped Transformer network containing channel-wise multi-scale feature fusion Transformer modules and spatial-wise global feature modeling Transformer modules specifically designed for UIE tasks, enhancing network attention to underwater image color channels and spatial regions. However, the self-attention mechanism in Transformers necessitates computing the relationships between each pixel and all other pixels, resulting in a quadratic computational complexity. For high-resolution underwater images, this computation can be excessively resource-intensive. Overcoming this limitation remains an important open challenge in the field of UIE for underwater applications.
\subsection{State Space Models (SSM)}
The SSM \cite{34,35,36} were originally developed in control theory \cite{37} and later introduced to deep learning due to their powerful capabilities in modeling sequential data. The SSM can effectively capture long-range dependencies in sequence data with linear complexity, drawing significant attention. Gu et al. \cite{34} first proposed a structured state space sequence model (S4) based on a new parameterization, proving it can be computed more efficiently while retaining theoretical advantages. Furthermore, Smith et al. \cite{36} introduced a new state space layer S5 utilizing an efficient parallel scanning mechanism. More recently, Gu et al. \cite{19} proposed the selective state space model (Mamba) with the advantages of fast inference and linear scalability with sequence length, outperforming Transformer on natural language tasks. Furthermore, Mamba has since been extended to computer vision including image classification \cite{20}, medical segmentation \cite{21}, and image recovery \cite{22,23}. However, existing Mamba-based methods for image recovery typically adopt global pooling to obtain channel attention, potentially losing spatial object information. Meanwhile, underwater images commonly lack red due to light absorption and scattering by water molecules, with increasing color distortion at depth. 

To address the limitations of high complexity in the Transformer and severe degradation in underwater images, we propose WaterMamba, a novel linear-complexity UIE method inspired by Mamba. WaterMamba effectively models underwater image pixel and color characteristics through a SCOSS block, better suited for UIE tasks while maintaining high model robustness.

\section{Methodology}

\begin{figure}
  \centering
\centerline{\includegraphics[height=9.6cm,width=13.9cm]{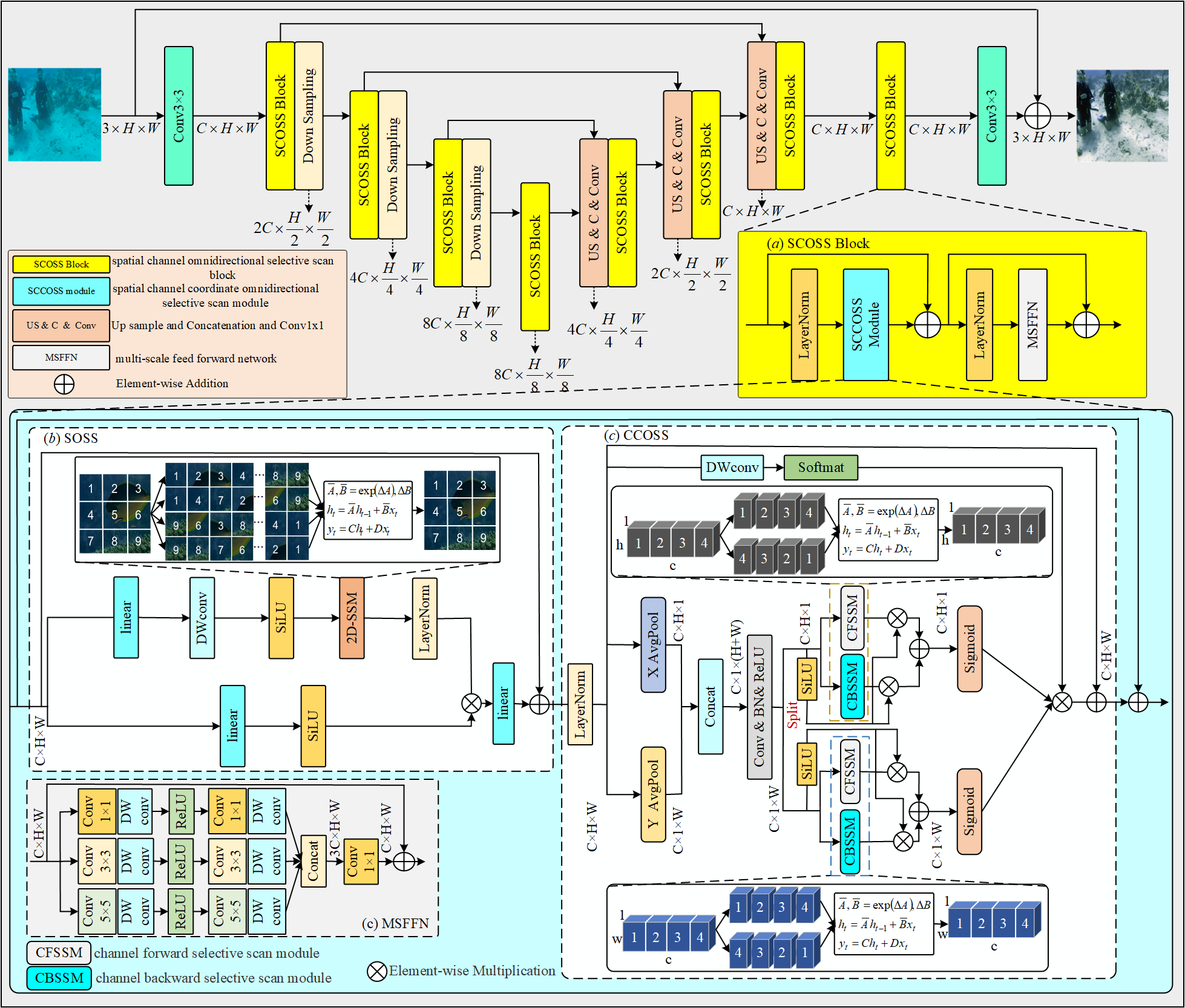}}\vspace{0.01cm}
  \caption{The architecture of the WaterMamba. (a) SCOSS; (b) SOSS; (c) CCOSS.}
  \label{fig2}
\end{figure}
In this section, the characteristics of underwater imaging are integrated with the latest advanced SSM, specifically Mamba, to introduce it into UIE. The preparation work of SSM is first described. Then, the overall network pipeline of WaterMamba is outlined, as shown in Fig. ~\ref{fig2}. Subsequently, the proposed SCOSS block is elaborated in detail, comprising the SCCOSS module and the MSFFN.
\subsection{Preliminaries}
The SSM is inspired by continuous linear time-invariant (LTI) systems and represents a class of sequence-to-sequence modeling systems with constant dynamic characteristics over time. It maps a one-dimensional function or sequence $x(t)\in R$ to an output sequence $y(t)\in R$ through an implicit latent state $h(t)\in R^N$. Therefore, the linear ordinary differential equation (ODE) form of the SSM can be defined as:
\begin{displaymath}
h^{\prime}(t)=Ah(t)+Bx(t),(1)  , y(t)=Ch(t)+Dx(t),(2)
\end{displaymath}
where $h(t)=R^N$ is the hidden state, $A=R^{N\times N}$, $B=R^{N}$ and $C=R^{N}$ are the parameters, when the state size is $N$, and $D=R^{1}$ represents the skip connection.

Subsequently, Eqs. (1) and (2) should be discretized. Specifically, the ODE in Eq. 1 can be discretized using the zero-order hold (ZOH) rule, which requires the continuous parameters $A$ and $B$ to be transformed into discrete parameters $\overline{A}$ and $\overline{B}$ using the time scale parameter $\Delta$. The discretization of Eqs. (1) and (2) can be defined as follows:
\begin{displaymath}
h^{\prime}\big(t\big)=\overline{A}h_{t-1}+\overline{B}x\big(t\big),(3) , y(t)=Ch_t+Dx_t,(4)
\end{displaymath}
\begin{displaymath}
\overline{A}=e^{\Delta A},(5) , \overline{B}=(\Delta A)^{-1}(e^{\Delta A}-I),(6)
\end{displaymath}
where $\Delta=R^{D}$ is the time scale parameter, and $B, C=R^{D\times N}$.

Compared to previous SSM and existing Mamba-based image restoration methods \cite{22,23}, the proposed spatial channel wise full Mamba leverages the selective mechanism introduced in Mamba \cite{19}. This allows it to capture pixel and channel features in underwater images, paying more attention to the loss feature in spatial and color channels, thereby significantly improving computational efficiency and accuracy.

\subsection{Overall Architecture}
The proposed WaterMamba adopts a U-shaped encoder-decoder structure, fully leveraging multi-scale features. The proposed WaterMamba mainly consists of an encoder, a decoder, and skip connection modules, as shown in Fig. ~\ref{fig2}. Given a low-quality underwater image $I=R^{3\times H \times W}$, the encoder part, $I$ is first passed through convolutional layers to obtain shallow features $F_1=R^{C\times H \times W}$. Then, the features $F_1$ are encoded through three SCOSS blocks and downsampling in the encoder. Each SCOSS block processes the features $F_2$, $F_3$, and $F_4$  respectively, which are then downsampled to sizes of $\frac H2\times\frac W2$, $\frac H4\times\frac W4$, and $\frac H8\times\frac W8$. The encoder, with its three SCOSS blocks and downsampling modules, progressively extracts multi-scale features of the image. By employing the linear complexity of the state space mechanism, it effectively captures long-range dependencies and inter-channel interaction information in the image. Next, at the end of the encoder, a bottleneck layer is designed to process $F_4$ and obtain $L$, where the bottleneck layer is a SCOSS block. Then, the feature $L$ is decoded through three SCOSS blocks and upsampling in the decoder. Each SCOSS block processes the features $D_1$, $D_2$, and $D_3$, where $D_1$ is the upsampling of $F_4$ and the first layer of the decoder to size $\frac H4\times\frac W4$, $D_2$ is the upsampling of $F_3$ and the second layer of the decoder to size $\frac H2\times\frac W2$, and $D_3$ is the upsampling of $F_2$ and the third layer of the decoder to size $H \times W$. Subsequently, $D_3$ is refined through the SCOSS block, and finally, through convolution, the feature $DR=R^{3\times H \times W}$ is obtained. Finally, the output image $FR=R^{3\times H \times W}$ is obtained by $FR=DR+I$. Overall, the WaterMamba network integrates SCOSS blocks, fully utilizing multi-scale feature information and global contextual relationships. While maintaining low complexity, it can generate high-quality underwater enhanced images.

\subsection{Spatial-Channel Omnidirectional Selective Scan Block}
Due to the significant absorption and scattering effects of water molecules on light, particularly in the longer wavelength range (such as red and infrared) \cite{5}, underwater images often lack red hues. As scene depth increases, the light in the water gradually diminishes, and colors become progressively distorted. In deep-water regions, images tend to exhibit a bluish-green tint, and details are lost more prominently. Therefore, considering the severe loss of detail and color distortion in underwater images, we have designed the SCOSS block to effectively model the correlation between pixels and channels in underwater images, with a greater focus on spatial and color channel information. As illustrated in Fig. ~\ref{fig2}, the SCOSS block comprises three key components: a spatial omnidirectional selective scan (SOSS) module, a channel coordinate omnidirectional selective scan (COOSS) module, and a multi-scale feedforward network (MSFFN).

\textbf{SOSS:} In terms of pixel spatial space, to maintain computational efficiency, transformer-based underwater enhancement methods often adopt shifted window attention \cite{29}, which limits modeling on the horizontal level of the entire image. However, underwater images often suffer from global losses. Inspired by the success of the Swin Transformer in long-range modeling with linear complexity, to better utilize two-dimensional spatial information, a two-dimensional selective scan module (2D-SSM) has been introduced into UIE. As shown in Fig. ~\ref{fig2}(b), the input feature $X=R^{C\times H \times W}$ goes through two parallel branches. In the first branch, feature channel expansion is performed through a linear layer, followed by a depth-wise convolution, SiLU activation function, and a 2D-SSM layer with Layer Normalization, defined as follows:
\begin{displaymath}
X_1=LN\big(2D-SSM\big(SiLU\big(DWConv\big(Linear\big(X\big)\big)\big)\big),(7)
\end{displaymath}
where DWConv represents the depth-wise convolution, and LN represents the layer normalization layer. The 2D-SSM layer scans the two-dimensional image features in four different directions: top-left to bottom-right, bottom-right to top-left, top-right to bottom-left, and bottom-left to top-right, flattening the two-dimensional image features into one-dimensional sequences. Subsequently, the long-range dependencies of each sequence are captured according to the discrete-state space equation in Eq. (3). Finally, all sequences are merged through summation, and a reshaping operation is performed to restore the two-dimensional structure. This method enables the adaptive capture of important pixel information in the image, enhancing the network's modeling capability for spatial details.

Then, in the second branch, feature channel expansion is first performed through a linear layer, followed by a SiLU activation function, which is defined as follows:
\begin{displaymath}
X_2=SiLU(Linear(X)),(8)
\end{displaymath}
Finally, the features from the two branches are aggregated through the Hadamard product. Lastly, the channels are projected to an output Xout with the same shape as the input, which is defined as follows:
\begin{displaymath}
X_{out}=Linear\big(X_1 ~\odot~ X_2\big)+X,(9)
\end{displaymath}
where $~\odot~$ representing the Hadamard product.

\textbf{COOSS:} In the channel aspect, existing image restoration methods based on Mamba typically use global max pooling/average pooling to compute channel attention. However, this method loses spatial feature of scene objects \cite{22,23}. Inspired by coordinate attention \cite{38}, in addition to introducing Mamba for channel attention, we also incorporate spatial attention, embedding positional information into channel attention. Similar to pixel scanning, we design a channel coordinate selective scan mechanism. With learnable channel scanning weights, the network can adaptively focus on channel features that are more important for the enhancement task, improving color perception.

As shown in Fig. ~\ref{fig2}(c), the input feature $Y=R^{C\times H \times W}$ is processed through global average pooling separately along the height and width dimensions, resulting in feature maps $Y_H=R^{C\times 1 \times W}$ and $Y_W=R^{C\times H \times 1}$ for the height and width directions, respectively. After applying average pooling along the width dimension, the features are mapped to the height dimension. Similarly, after applying average pooling along the height dimension, the features are mapped to the width dimension. This process is defined as follows:
\begin{displaymath}
Y_H=X_{A\nu gPool}(Y),Y_W=X_{A\nu gPool}(Y),(10)
\end{displaymath}
After applying average pooling along the height dimension, the features are mapped to the width dimension. The two parallel stages are then merged by transposing the width and height to the same dimension, followed by stacking to combine the width and height features. Subsequently, a convolution operation, normalization, and activation function are applied to obtain the feature $Y_1=R^{C\times 1 \times (H+W)}$, which is defined as follows:
\begin{displaymath}
X_{out}=Y_1=ReLU(BN(Conv(Concat(Y_H,Y_W)))),(11)
\end{displaymath}
Afterwards, the process is split into two parallel stages again, separating the width and height into $Y_{H1}=R^{C\times 1 \times W}$ and $Y_{W1}=R^{C\times H \times 1}$, respectively, which are defined as follows:
\begin{displaymath}
Y_{H1},Y_{W1}=Split(Y_1),(12)
\end{displaymath}
Then, after channel selective scan using the Mamba block, a sigmoid activation function is applied to obtain the attention map $Y_{H2}$ along the height dimension, which is defined as follows:
\begin{displaymath}
Y_{H2}=Sigmoid\left(CFSSM\left(Y_{H1}\right)\odot SiLU\left(Y_{H1}\right)+CBSSM\left(Y_{H1}\right) \odot SiLU\left(Y_{H1}\right)\right),(13)
\end{displaymath}
where, CFSSM is the channel selective forward scan module and CBSSM is the channel backward selective scan module.

Subsequently, channel-selective scanning is performed using the Mamba block, followed by a sigmoid activation function to obtain the attention map $Y_{W2}$ along the width dimension, which is defined as follows:
\begin{displaymath}
Y_{W2}=Sigmoid\left(CFSSM\left(Y_{W1}\right) \odot SiLU\left(Y_{W1}\right)+CBSSM\left(Y_{W1}\right) \odot SiLU\left(Y_{W1}\right)\right),(14)
\end{displaymath}
Finally, the Hadamard product of the attention maps $Y_{H2}$ and $Y_{W2}$ with the features obtained by applying 1×1 depth-wise convolution and softmax activation to $Y$ is added to $Y$. This process is defined as follows:
\begin{displaymath}
Y_{out}=Y_{H2} \odot Y_{W2}  \odot Y  \odot Softmat(DWConv(Y))+Y,(15)
\end{displaymath}
\textbf{MSFFN:} As shown in Fig. ~\ref{fig2}(c), considering that using only the Mamba module alone may lead to pattern collapse, the features are subjected to layer normalization after being modeled by the SCOSS module. Subsequently, the features are aggregated through a MSFFN. The MSFFN employs ReLU activation, ensuring flexibility in feature aggregation. The MSFFN consists of three branches, and the first branch can be defined as follows:
\begin{displaymath}
Z_1=D_W^1C^1\left(ReLU{\left(D_W^1C^1(LN(Z))\right)}\right),(16)
\end{displaymath}
where, $D_W^1$ denotes a 1×1 depth-wise convolution, while $C^1$ represents a 1×1 convolution.
The second branch can be defined as follows:
\begin{displaymath}
Z_3=D_W^3C^3\left(ReLU{\left(D_W^3C^3(LN(Z))\right)}\right),(17)
\end{displaymath}
where, $D_W^3$ denotes a 3×3 depth-wise convolution, while $C^3$ represents a 3×3 convolution.

The third branch can be defined as follows:
\begin{displaymath}
Z_5=D_W^5C^5\left(ReLU{\left(D_W^5C^5(LN(Z))\right)}\right),(18)
\end{displaymath}
where, $D_W^5$ denotes a 5×5 depth-wise convolution, while $C^5$ represents a 5×5 convolution.

Finally, the SCOSS block can be represented as follows:
\begin{displaymath}
Z_{out}=MSFFN\big(CCOSS\big(SOSS\big(X\big)\big)+X\big)+Z,(19)
\end{displaymath}
Through the SCOSS module, the WaterMamba network is able to learn spatial and color features of underwater images comprehensively and adaptively. Compared to traditional methods like global pooling, the SCOSS module can better capture important local information, leading to the generation of more natural and clear underwater enhanced images. The linear complexity of the SCOSS module also ensures that the entire WaterMamba network achieves excellent UIE performance while maintaining low computational costs.

\section{Experiments and Analysis}
\subsection{Implementation details}
The proposed WaterMamba was implemented using the PyTorch 2.0.0 framework and was trained and tested on an NVIDIA RTX 3090 GPU. The network was trained end-to-end using the AdamW \cite{39} optimizer with a learning rate of 1e-4 and $\beta_1 = 0.9, \beta_2 = 0.999$. All images were resized to a fixed size of 256x256 pixels. A training batch size of 2 was used, and the network was trained for 750 epochs. The learning rate was dynamically adjusted using a cosine annealing strategy \cite{28}.
\subsection{Datasets and Evaluation Metrics}
\textbf{Dataset:} The experiment validation was conducted using four publicly available underwater image datasets, including UIEB\cite{10}, SQUID\cite{8}, UCCS\cite{40} and UCIOD\cite{15}. The UIEB dataset consists of 890 real underwater images and 60 real images without reference (C60). The UIEB dataset was divided into a training set of 800 image pairs and a validation set (R90) of 90 image pairs, with each pair consisting of a degraded underwater image and its corresponding clear reference image. The UCIOD dataset is a real underwater dataset, where each image pair is from the same underwater video. It includes 50 different scenes with similar degradation levels, and 1000 images were randomly selected as the test set (U100). The UCCS dataset contains real images from three scenes: 100 blue-shifted, 100 green-shifted, and 100 blue-green-shifted images. The SQUID dataset consists of 57 pairs of stereo underwater images captured from four different diving locations, including coral reefs and shipwrecks in the Red Sea and rocky reefs in the Mediterranean. 16 images were selected as the test set (S16). The R90, C60, U100, UCCS, and S16 datasets were chosen to evaluate the effectiveness of the proposed WaterMamba method. These datasets cover various underwater scenes and imaging conditions, such as coral reefs, underwater organisms, and underwater terrains.

\textbf{Evaluation Metrics:} Four objective evaluation metrics were used to comprehensively assess the performance of the WaterMamba method: Peak signal-to-noise ratio (PSNR) \cite{41} : Reflects the ratio of the image signal to noise and is used to evaluate the overall quality of the image. Structural similarity index (SSIM) \cite{42}: Measures the structural similarity of the image, which closely aligns with human perception. Underwater image quality measure (UIQM) \cite{43}: A dedicated quality evaluation metric for underwater images, encompassing color, contrast, and clarity. Underwater color image quality evaluation (UCIQE) \cite{44}: Evaluates the smoothness, clarity, and contrast of color quantization in underwater images.
\subsection{Comparision with SOTA Methods}
\subsubsection{Compared Methods}
To validate the effectiveness of the proposed WaterMamba method, it was compared with 10 SOTA methods for UIE. These methods include 6 CNN-based methods (Ucolor \cite{13}, PUIE-Net (MC) \cite{14}, PUIE-Net (MP) \cite{14}, PUGAN \cite{25}, MFEF \cite{26} and Semi-UIR \cite{27}) and 4 Transformer-based methods (Restormer  \cite{28}, URSCT \cite{29}, X-CAUNET \cite{5}, Convformer \cite{33}). The results of these 10 methods were obtained by using the code released by the corresponding authors. Comparing WaterMamba with these SOTA methods, it is aimed to demonstrate the superiority of WaterMamba in enhancing underwater images.
\subsubsection{Qualitative Comparisons}

\begin{figure*}[!t]    
	\centering
	\subfigure{
		\begin{minipage}[b]{0.058\linewidth}
			\centerline{\includegraphics[width=1.0cm,height=1.0cm]{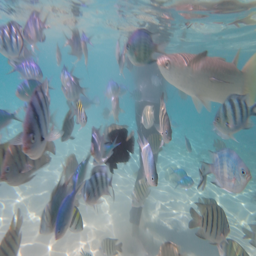}}\vspace{0.01cm}
			\centerline{\includegraphics[width=1.0cm,height=1.0cm]{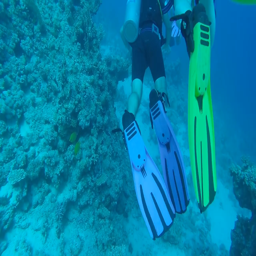}}\vspace{0.01cm} 
			\centerline{\includegraphics[width=1.0cm,height=1.0cm]{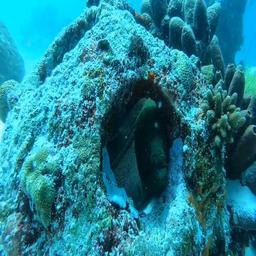}}\vspace{0.01cm}
			\centerline{\includegraphics[width=1.0cm,height=1.0cm]{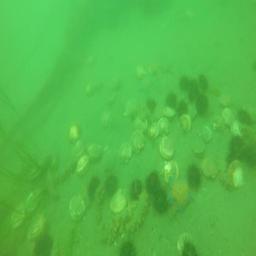}}\vspace{0.01cm} 

			\centerline{(a)}
		\end{minipage}
	}
	\subfigure{
		\begin{minipage}[b]{0.058\linewidth}
			\centerline{\includegraphics[width=1.0cm,height=1.0cm]{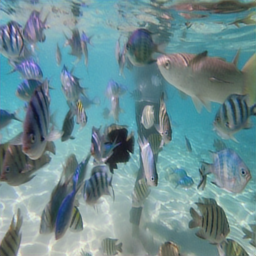}}\vspace{0.01cm}
			\centerline{\includegraphics[width=1.0cm,height=1.0cm]{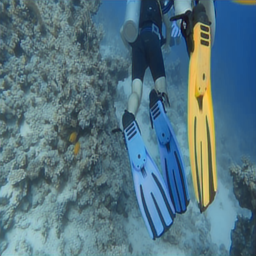}}\vspace{0.01cm} 
			\centerline{\includegraphics[width=1.0cm,height=1.0cm]{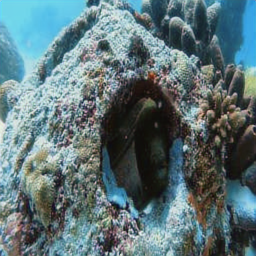}}\vspace{0.01cm}
			\centerline{\includegraphics[width=1.0cm,height=1.0cm]{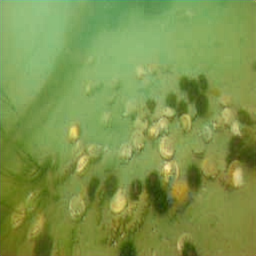}}\vspace{0.01cm} 

			\centerline{(b)}
		\end{minipage}
	}
	\subfigure{
		\begin{minipage}[b]{0.058\linewidth}
			\centerline{\includegraphics[width=1.0cm,height=1.0cm]{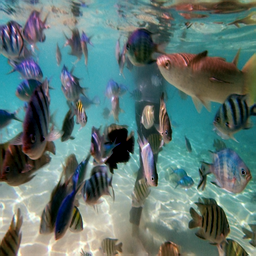}}\vspace{0.01cm}
			\centerline{\includegraphics[width=1.0cm,height=1.0cm]{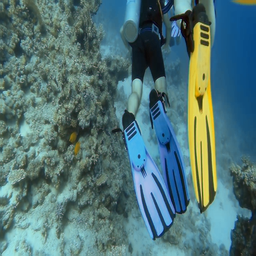}}\vspace{0.01cm} 
			\centerline{\includegraphics[width=1.0cm,height=1.0cm]{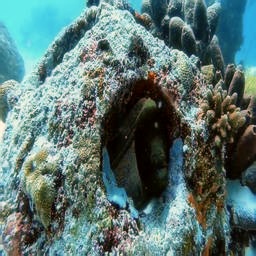}}\vspace{0.01cm}
			\centerline{\includegraphics[width=1.0cm,height=1.0cm]{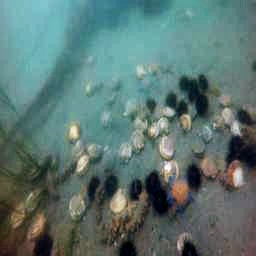}}\vspace{0.01cm}

			\centerline{(c)}
		\end{minipage}
	}
	\subfigure{
		\begin{minipage}[b]{0.058\linewidth}
			\centerline{\includegraphics[width=1.0cm,height=1.0cm]{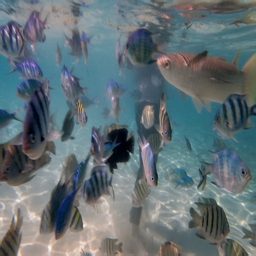}}\vspace{0.01cm}
			\centerline{\includegraphics[width=1.0cm,height=1.0cm]{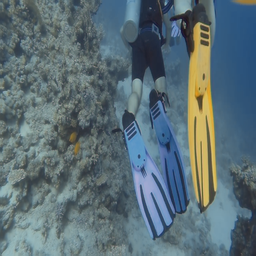}}\vspace{0.01cm} 
			\centerline{\includegraphics[width=1.0cm,height=1.0cm]{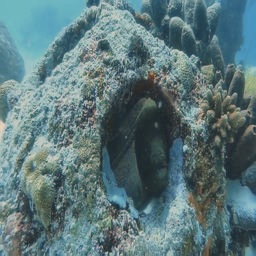}}\vspace{0.01cm}
			\centerline{\includegraphics[width=1.0cm,height=1.0cm]{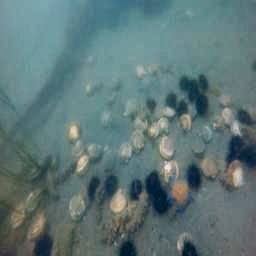}}\vspace{0.01cm}

			\centerline{(d)}
		\end{minipage}
	}
	\subfigure{
		\begin{minipage}[b]{0.058\linewidth}
			\centerline{\includegraphics[width=1.0cm,height=1.0cm]{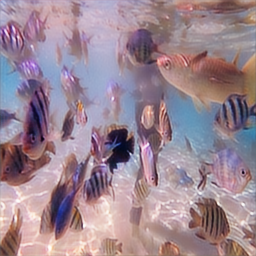}}\vspace{0.01cm}
			\centerline{\includegraphics[width=1.0cm,height=1.0cm]{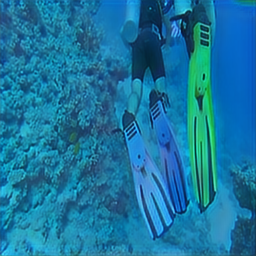}}\vspace{0.01cm} 
			\centerline{\includegraphics[width=1.0cm,height=1.0cm]{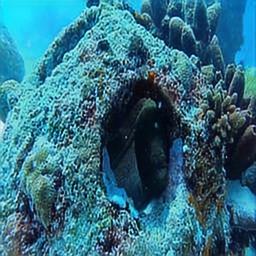}}\vspace{0.01cm}
			\centerline{\includegraphics[width=1.0cm,height=1.0cm]{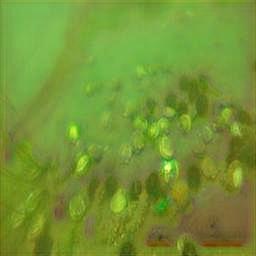}}\vspace{0.01cm}

			\centerline{(e)}
		\end{minipage}
	}
	\subfigure{
		\begin{minipage}[b]{0.058\linewidth}
			\centerline{\includegraphics[width=1.0cm,height=1.0cm]{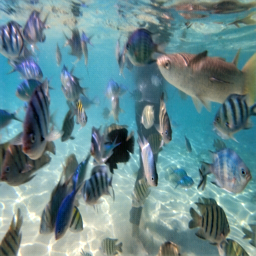}}\vspace{0.01cm}
			\centerline{\includegraphics[width=1.0cm,height=1.0cm]{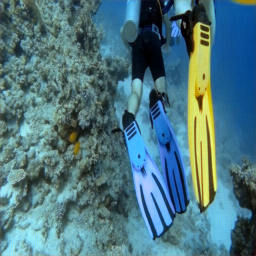}}\vspace{0.01cm} 
			\centerline{\includegraphics[width=1.0cm,height=1.0cm]{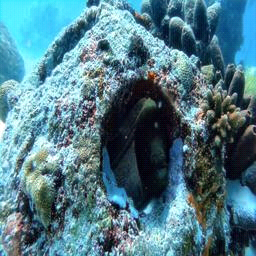}}\vspace{0.01cm}
			\centerline{\includegraphics[width=1.0cm,height=1.0cm]{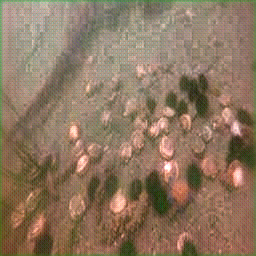}}\vspace{0.01cm}

			\centerline{(f)}
		\end{minipage}
	}
	\subfigure{
		\begin{minipage}[b]{0.058\linewidth}
			\centerline{\includegraphics[width=1.0cm,height=1.0cm]{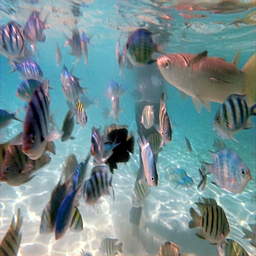}}\vspace{0.01cm}
			\centerline{\includegraphics[width=1.0cm,height=1.0cm]{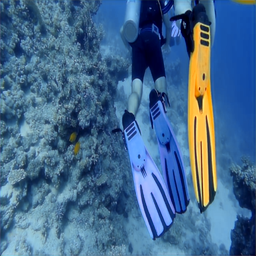}}\vspace{0.01cm} 
			\centerline{\includegraphics[width=1.0cm,height=1.0cm]{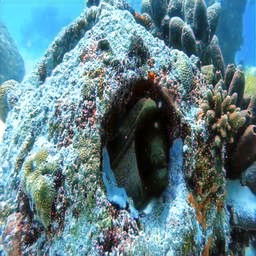}}\vspace{0.01cm}
			\centerline{\includegraphics[width=1.0cm,height=1.0cm]{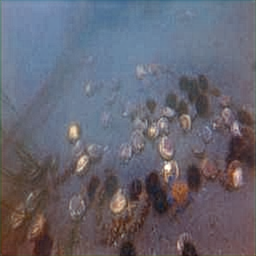}}\vspace{0.01cm}

			\centerline{(g)}
		\end{minipage}
	}
	\subfigure{
		\begin{minipage}[b]{0.058\linewidth}
			\centerline{\includegraphics[width=1.0cm,height=1.0cm]{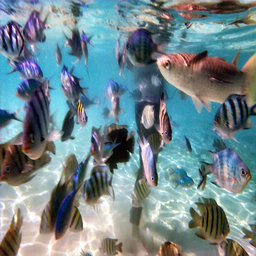}}\vspace{0.01cm}
			\centerline{\includegraphics[width=1.0cm,height=1.0cm]{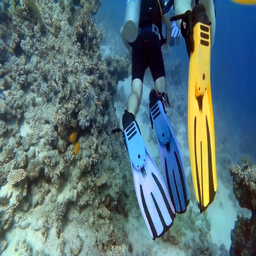}}\vspace{0.01cm} 
			\centerline{\includegraphics[width=1.0cm,height=1.0cm]{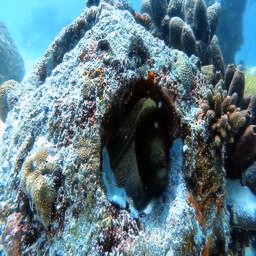}}\vspace{0.01cm}
			\centerline{\includegraphics[width=1.0cm,height=1.0cm]{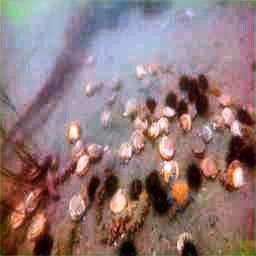}}\vspace{0.01cm}

			\centerline{(h)}
		\end{minipage}
	}
	\subfigure{
		\begin{minipage}[b]{0.058\linewidth}
			\centerline{\includegraphics[width=1.0cm,height=1.0cm]{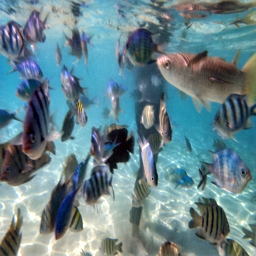}}\vspace{0.01cm}
			\centerline{\includegraphics[width=1.0cm,height=1.0cm]{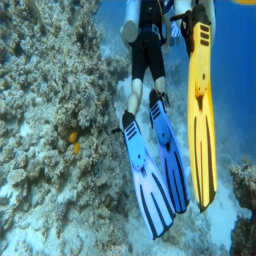}}\vspace{0.01cm} 
			\centerline{\includegraphics[width=1.0cm,height=1.0cm]{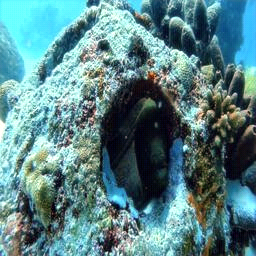}}\vspace{0.01cm}
			\centerline{\includegraphics[width=1.0cm,height=1.0cm]{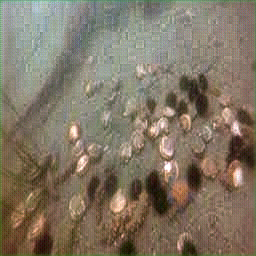}}\vspace{0.01cm} 
      
			\centerline{(i)}
		\end{minipage}
	}
	\subfigure{
		\begin{minipage}[b]{0.058\linewidth}
			\centerline{\includegraphics[width=1.0cm,height=1.0cm]{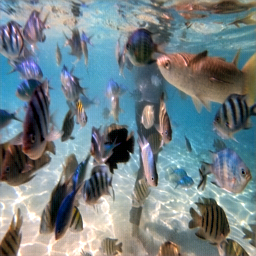}}\vspace{0.01cm}
			\centerline{\includegraphics[width=1.0cm,height=1.0cm]{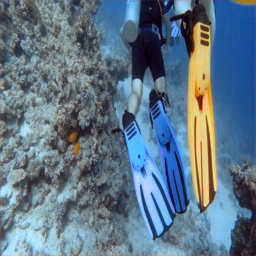}}\vspace{0.01cm} 
			\centerline{\includegraphics[width=1.0cm,height=1.0cm]{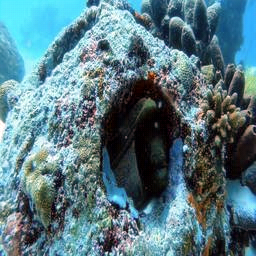}}\vspace{0.01cm}
			\centerline{\includegraphics[width=1.0cm,height=1.0cm]{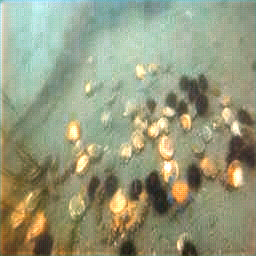}}\vspace{0.01cm}

			\centerline{(j)}
		\end{minipage}
	}
	\subfigure{
		\begin{minipage}[b]{0.058\linewidth}
			\centerline{\includegraphics[width=1.0cm,height=1.0cm]{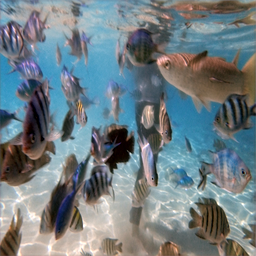}}\vspace{0.01cm}
			\centerline{\includegraphics[width=1.0cm,height=1.0cm]{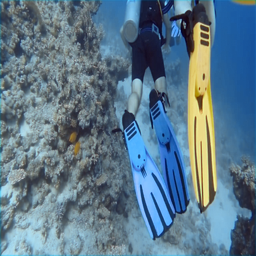}}\vspace{0.01cm} 
			\centerline{\includegraphics[width=1.0cm,height=1.0cm]{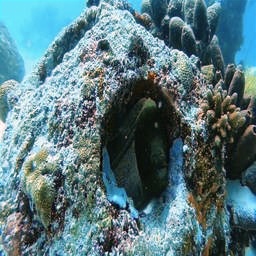}}\vspace{0.01cm}
			\centerline{\includegraphics[width=1.0cm,height=1.0cm]{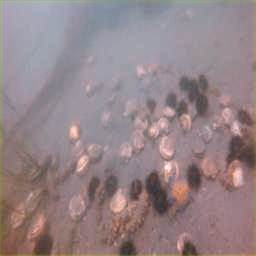}}\vspace{0.01cm}

			\centerline{(k)}
		\end{minipage}
	}
	\subfigure{
		\begin{minipage}[b]{0.058\linewidth}
			\centerline{\includegraphics[width=1.0cm,height=1.0cm]{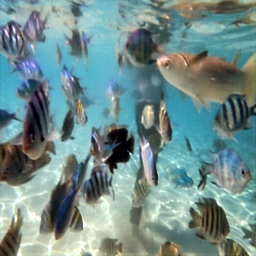}}\vspace{0.01cm}
			\centerline{\includegraphics[width=1.0cm,height=1.0cm]{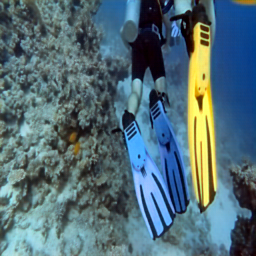}}\vspace{0.01cm} 
			\centerline{\includegraphics[width=1.0cm,height=1.0cm]{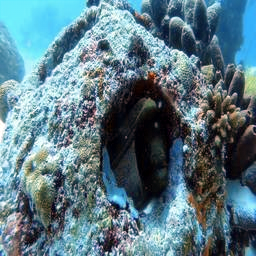}}\vspace{0.01cm}
			\centerline{\includegraphics[width=1.0cm,height=1.0cm]{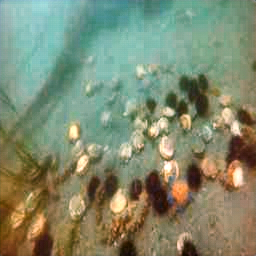}}\vspace{0.01cm}

			\centerline{(l)}
		\end{minipage}
	}
	\subfigure{
		\begin{minipage}[b]{0.058\linewidth}
			\centerline{\includegraphics[width=1.0cm,height=1.0cm]{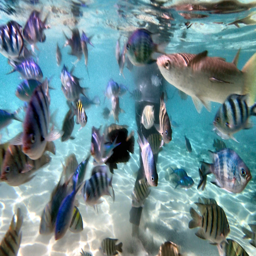}}\vspace{0.01cm}
			\centerline{\includegraphics[width=1.0cm,height=1.0cm]{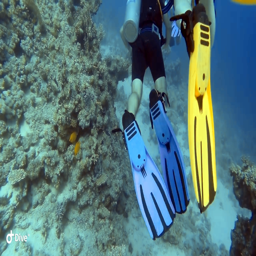}}\vspace{0.01cm} 
			\centerline{\includegraphics[width=1.0cm,height=1.0cm]{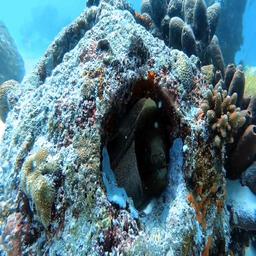}}\vspace{0.01cm}
			\centerline{\includegraphics[width=1.0cm,height=1.0cm]{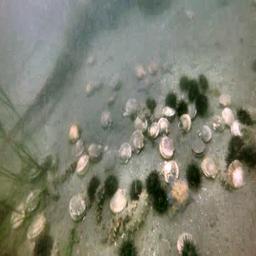}}\vspace{0.01cm}

			\centerline{(m)}
		\end{minipage}
	}

  \caption{The visual contrasts from top to bottom for the different datasets are UIEB(R90 and C60)\cite{10} and U100\cite{15} in each row. (a) raw; (b) Ucolor\cite{13}; (c) PUIE-Net(MC)\cite{14}; (d) PUIE-Net(MP)\cite{14}; (e) PUGAN\cite{25}; (f) MFEF\cite{26}; (g) Semi-UIR\cite{27}; (h) URSCT\cite{29}; (i) Restormer\cite{28}; (j) Convformer\cite{33} (k) X-CAUNET\cite{5}; (l) WaterMamba; (m) reference.
}
  \label{fig3}
\end{figure*}

\begin{figure*}[!t]    
	\centering
	\subfigure{
		\begin{minipage}[b]{0.058\linewidth}

			\centerline{\includegraphics[width=1.0cm,height=1.0cm]{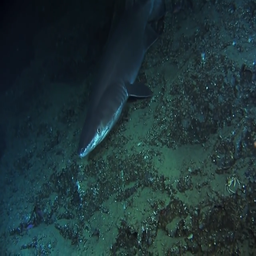}}\vspace{0.01cm}
			\centerline{\includegraphics[width=1.0cm,height=1.0cm]{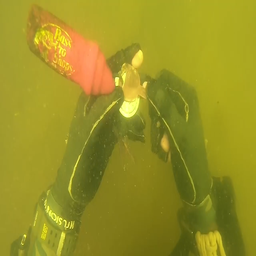}}\vspace{0.01cm} 
			\centerline{\includegraphics[width=1.0cm,height=1.0cm]{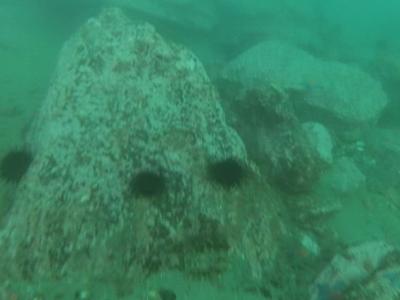}}\vspace{0.01cm}
			\centerline{\includegraphics[width=1.0cm,height=1.0cm]{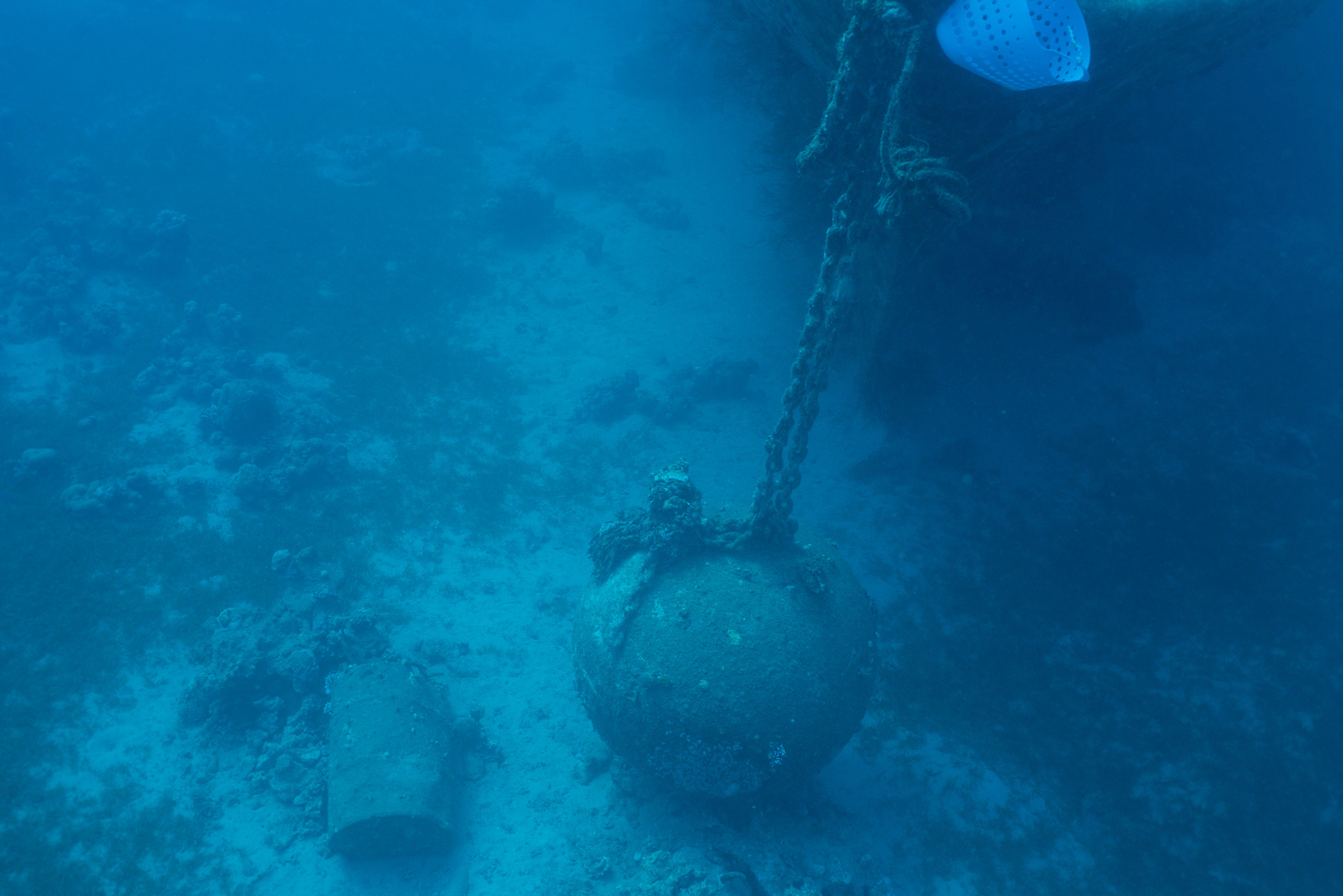}}\vspace{0.01cm} 
			\centerline{(a)}
		\end{minipage}
	}
	\subfigure{
		\begin{minipage}[b]{0.058\linewidth}

			\centerline{\includegraphics[width=1.0cm,height=1.0cm]{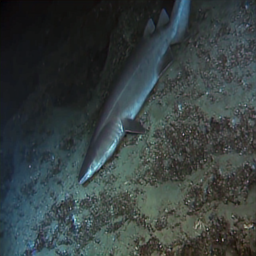}}\vspace{0.01cm}
			\centerline{\includegraphics[width=1.0cm,height=1.0cm]{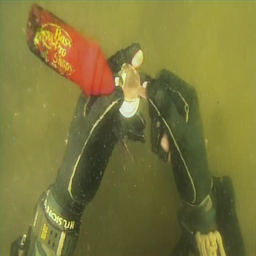}}\vspace{0.01cm} 
			\centerline{\includegraphics[width=1.0cm,height=1.0cm]{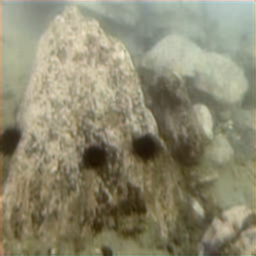}}\vspace{0.01cm}
			\centerline{\includegraphics[width=1.0cm,height=1.0cm]{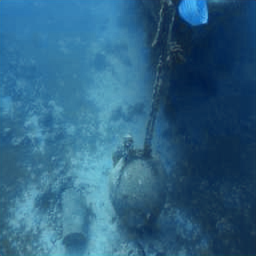}}\vspace{0.01cm} 
			\centerline{(b)}
		\end{minipage}
	}
	\subfigure{
		\begin{minipage}[b]{0.058\linewidth}

			\centerline{\includegraphics[width=1.0cm,height=1.0cm]{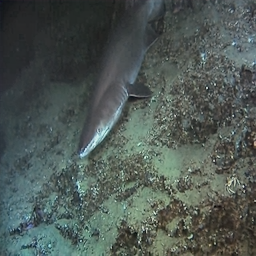}}\vspace{0.01cm}
			\centerline{\includegraphics[width=1.0cm,height=1.0cm]{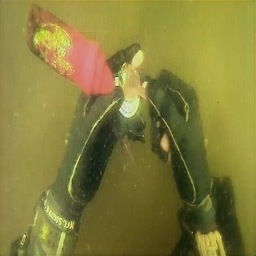}}\vspace{0.01cm} 
			\centerline{\includegraphics[width=1.0cm,height=1.0cm]{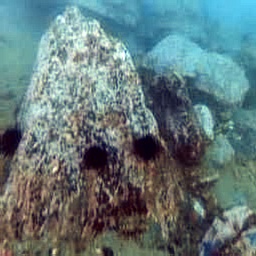}}\vspace{0.01cm}
			\centerline{\includegraphics[width=1.0cm,height=1.0cm]{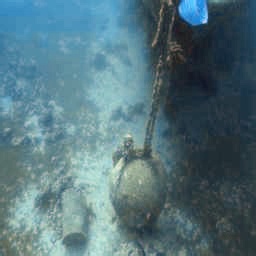}}\vspace{0.01cm}   
          
			\centerline{(c)}
		\end{minipage}
	}
	\subfigure{
		\begin{minipage}[b]{0.058\linewidth}

			\centerline{\includegraphics[width=1.0cm,height=1.0cm]{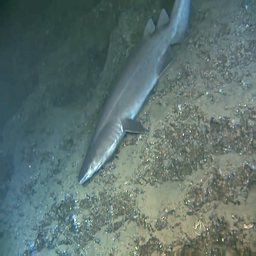}}\vspace{0.01cm}
			\centerline{\includegraphics[width=1.0cm,height=1.0cm]{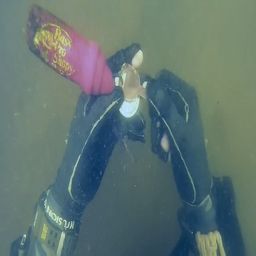}}\vspace{0.01cm} 
			\centerline{\includegraphics[width=1.0cm,height=1.0cm]{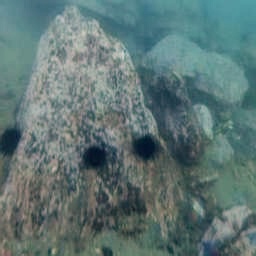}}\vspace{0.01cm}
			\centerline{\includegraphics[width=1.0cm,height=1.0cm]{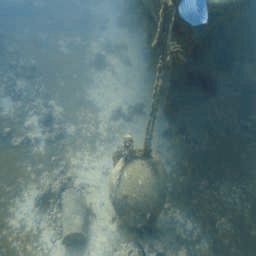}}\vspace{0.01cm} 
      
			\centerline{(d)}
		\end{minipage}
	}
	\subfigure{
		\begin{minipage}[b]{0.058\linewidth}

			\centerline{\includegraphics[width=1.0cm,height=1.0cm]{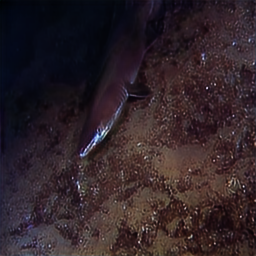}}\vspace{0.01cm}
			\centerline{\includegraphics[width=1.0cm,height=1.0cm]{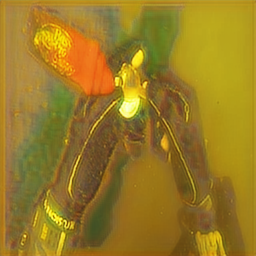}}\vspace{0.01cm} 
			\centerline{\includegraphics[width=1.0cm,height=1.0cm]{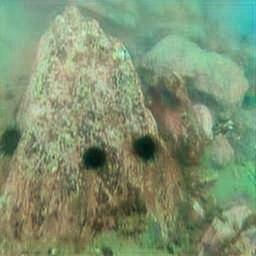}}\vspace{0.01cm}
			\centerline{\includegraphics[width=1.0cm,height=1.0cm]{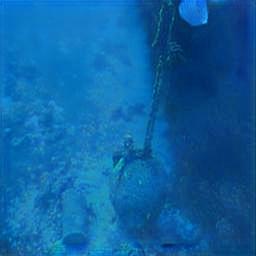}}\vspace{0.01cm} 
         
			\centerline{(e)}
		\end{minipage}
	}
	\subfigure{
		\begin{minipage}[b]{0.058\linewidth}

			\centerline{\includegraphics[width=1.0cm,height=1.0cm]{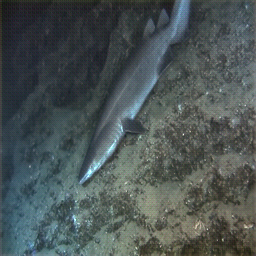}}\vspace{0.01cm}
			\centerline{\includegraphics[width=1.0cm,height=1.0cm]{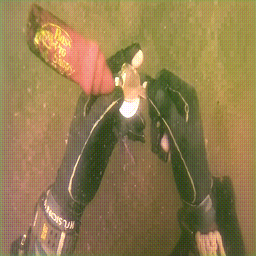}}\vspace{0.01cm} 
			\centerline{\includegraphics[width=1.0cm,height=1.0cm]{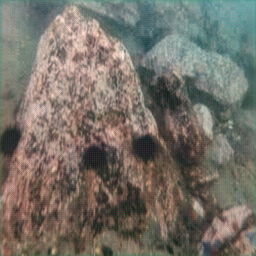}}\vspace{0.01cm}
			\centerline{\includegraphics[width=1.0cm,height=1.0cm]{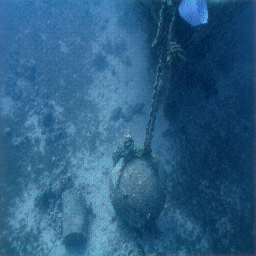}}\vspace{0.01cm} 
 
			\centerline{(f)}
		\end{minipage}
	}
	\subfigure{
		\begin{minipage}[b]{0.058\linewidth}

			\centerline{\includegraphics[width=1.0cm,height=1.0cm]{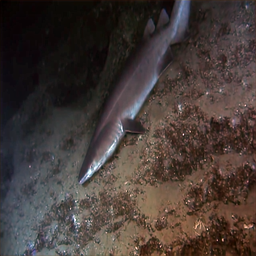}}\vspace{0.01cm}
			\centerline{\includegraphics[width=1.0cm,height=1.0cm]{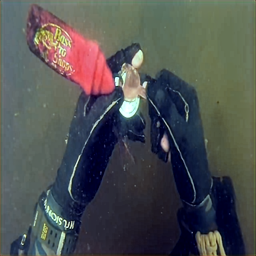}}\vspace{0.01cm} 
			\centerline{\includegraphics[width=1.0cm,height=1.0cm]{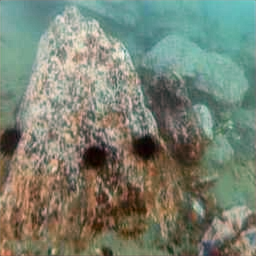}}\vspace{0.01cm}
			\centerline{\includegraphics[width=1.0cm,height=1.0cm]{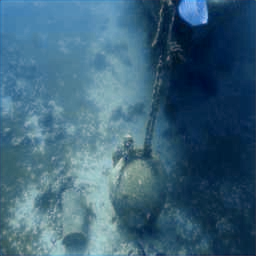}}\vspace{0.01cm}    
 
			\centerline{(g)}
		\end{minipage}
	}
	\subfigure{
		\begin{minipage}[b]{0.058\linewidth}

			\centerline{\includegraphics[width=1.0cm,height=1.0cm]{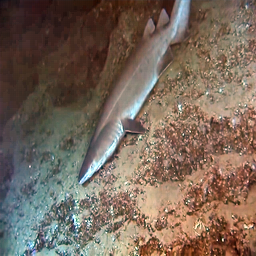}}\vspace{0.01cm}
			\centerline{\includegraphics[width=1.0cm,height=1.0cm]{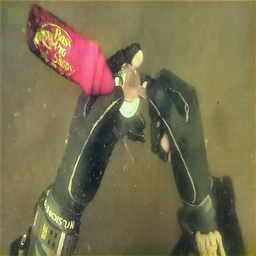}}\vspace{0.01cm} 
			\centerline{\includegraphics[width=1.0cm,height=1.0cm]{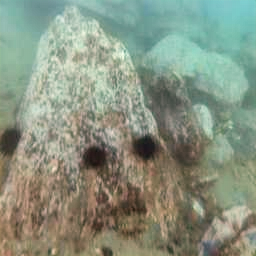}}\vspace{0.01cm}
			\centerline{\includegraphics[width=1.0cm,height=1.0cm]{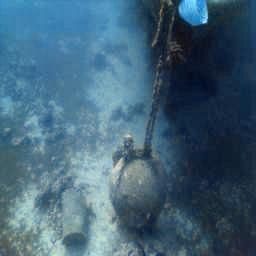}}\vspace{0.01cm}      
          
			\centerline{(h)}
		\end{minipage}
	}
	\subfigure{
		\begin{minipage}[b]{0.058\linewidth}

			\centerline{\includegraphics[width=1.0cm,height=1.0cm]{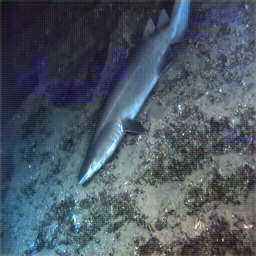}}\vspace{0.01cm}
			\centerline{\includegraphics[width=1.0cm,height=1.0cm]{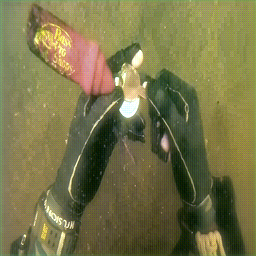}}\vspace{0.01cm} 
			\centerline{\includegraphics[width=1.0cm,height=1.0cm]{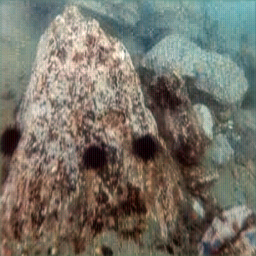}}\vspace{0.01cm}
			\centerline{\includegraphics[width=1.0cm,height=1.0cm]{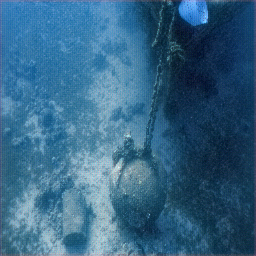}}\vspace{0.01cm}       
			\centerline{(i)}
		\end{minipage}
	}
	\subfigure{
		\begin{minipage}[b]{0.058\linewidth}

			\centerline{\includegraphics[width=1.0cm,height=1.0cm]{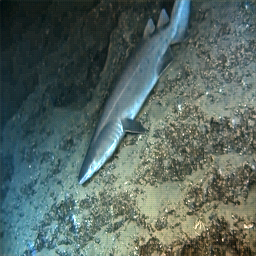}}\vspace{0.01cm}
			\centerline{\includegraphics[width=1.0cm,height=1.0cm]{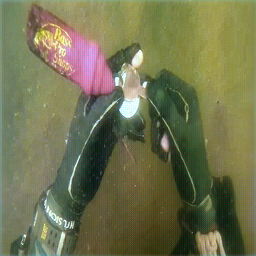}}\vspace{0.01cm} 
			\centerline{\includegraphics[width=1.0cm,height=1.0cm]{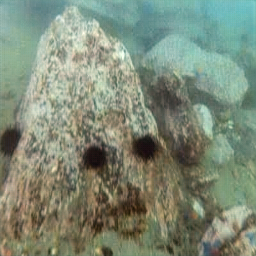}}\vspace{0.01cm}
			\centerline{\includegraphics[width=1.0cm,height=1.0cm]{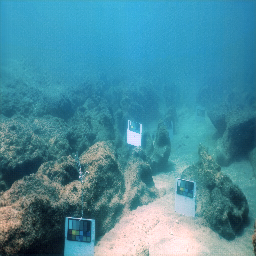}}\vspace{0.01cm}    
   
			\centerline{(j)}
		\end{minipage}
	}
	\subfigure{
		\begin{minipage}[b]{0.058\linewidth}
 
			\centerline{\includegraphics[width=1.0cm,height=1.0cm]{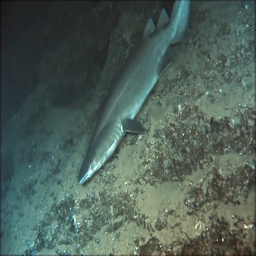}}\vspace{0.01cm}
			\centerline{\includegraphics[width=1.0cm,height=1.0cm]{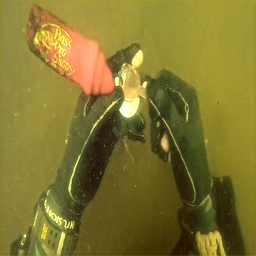}}\vspace{0.01cm} 
			\centerline{\includegraphics[width=1.0cm,height=1.0cm]{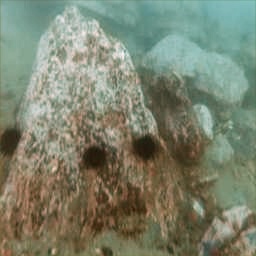}}\vspace{0.01cm}
			\centerline{\includegraphics[width=1.0cm,height=1.0cm]{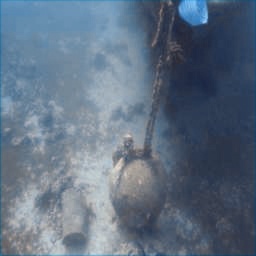}}\vspace{0.01cm}  
     
			\centerline{(k)}
		\end{minipage}
	}
	\subfigure{
		\begin{minipage}[b]{0.058\linewidth}

			\centerline{\includegraphics[width=1.0cm,height=1.0cm]{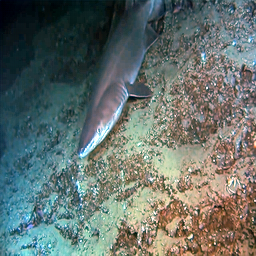}}\vspace{0.01cm}
			\centerline{\includegraphics[width=1.0cm,height=1.0cm]{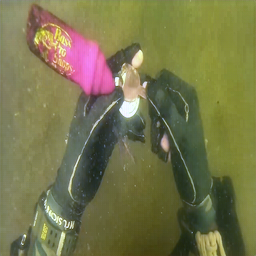}}\vspace{0.01cm} 
			\centerline{\includegraphics[width=1.0cm,height=1.0cm]{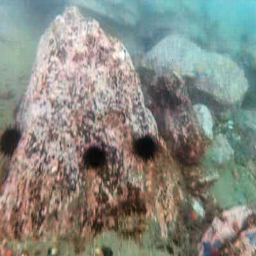}}\vspace{0.01cm}
			\centerline{\includegraphics[width=1.0cm,height=1.0cm]{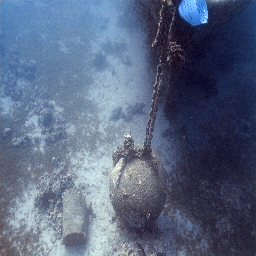}}\vspace{0.01cm}

			\centerline{(l)}
		\end{minipage}
	}

  \caption{The visual contrasts from top to bottom for the different datasets are UCCS\cite{40}, and SQUID\cite{8} in each row. (a) raw; (b) Ucolor\cite{13}; (c) PUIE-Net(MC)\cite{14}; (d) PUIE-Net(MP)\cite{14}; (e) PUGAN\cite{25}; (f) MFEF\cite{26}; (g) Semi-UIR\cite{27}; (h) URSCT\cite{29}; (i) Restormer\cite{28}; (j) Convformer\cite{33} (k) X-CAUNET\cite{5}; (l) WaterMamba.
}
  \label{fig4}
\end{figure*}

The visual comparison of our WaterMamba method with the other 10 SOTA methods is shown in Figs. ~\ref{fig3} and ~\ref{fig4}. We selected representative scenes from the four datasets, which cover a wide range of real underwater scenes. From Figs. ~\ref{fig3} and ~\ref{fig4}, it can be observed that Ucolor and PUGAN have certain limitations in restoring the details and color information of underwater images. The generated images have blurred details and color distortion. In contrast, WaterMamba not only effectively restores the clarity and contrast of the images but also preserves and enhances the color information, resulting in a more natural and clear visual effect. Particularly in complex underwater coral reefs and underwater terrain scenes, the superiority of WaterMamba is more prominent.
\subsubsection{Qualitative Comparisons}

\begin{table}[h]
	\caption{
 Quantitative method comparisons across \textbf{UIEB}, \textbf{UCIOD}, \textbf{UCCS}, and \textbf{SQUID} datasets, and parameter size. The performance and FLOPs are measured on an image size of 256 × 256. Top performances highlighted: best in Red,
second in blue.
	}
	\label{tab1}
	\centering
	\fontsize{7}{10.8}\selectfont
	\setlength{\tabcolsep}{1.080mm}{
	\begin{tabular}{ccccccccccccc}
		\hline
		 \multirow{2}{*}{ \textbf{Method}}& \multicolumn{2}{c}{\textbf{R90}} & \multicolumn{2}{c}{\textbf{U100}}& \multicolumn{2}{c}{\textbf{C60}}& \multicolumn{2}{c}{\textbf{UCCS}}& \multicolumn{2}{c}{\textbf{SQUID}}&\multirow{2}{*}{ \textbf{Params}}&\multirow{2}{*}{ \textbf{FLOPs}}
\\&PSNR&SSIM &PSNR&SSIM &UIQM&UCIQE &UIQM&UCIQE &UIQM&UCIQE \\
		\hline

		Ucolor (TIP 21)\cite{13}&21.093  &0.872 &20.483&  0.821 &2.482&  0.553 &3.019 &0.550 &2.042 &0.514 &157.4M &34.68G \\
	
		PUIE-Net (MC)(ECCV 22)\cite{14}&21.382 &0.882 &19.154 &0.819 &2.521 &0.558 &3.003 &0.536 &2.194 &0.495 &1.41M &30.09G \\

		PUIE-Net (MP)(ECCV 22)\cite{14}&19.257 &0.828 &17.019 &0.762 &2.398 &0.513 &2.758 &0.489 &2.054 &0.447 &1.41M &30.09G \\

		PUGAN (TIP 23)\cite{25}&22.653 &0.902 &17.517 &0.778 &2.652 &0.566 &2.977 &0.536 &2.225 &0.536 &95.66M &72.05G  \\

		MFEF (EAAI 23)\cite{26}&23.389 &0.918 &20.868 &0.816 &2.652 &0.566 &2.977 &0.556 &2.344 &0.545 &61.86M &26.52G \\
	
		Semi-UIR (CVPR 23)\cite{27}&\textcolor{blue}{24.119} &0.908 &\textcolor{blue}{21.318} &0.834 &2.667 &\textcolor{blue}{0.574} &\textcolor{red}{3.079} &0.554 &\textcolor{blue}{2.483} &0.556 &1.65M &36.44G \\

		URSCT (TGRS 22)\cite{29}	&23.754 &\textcolor{blue}{0.919} &21.252 &0.827 &2.642 &0.543 &2.947 &0.544 &2.226 &0.534 &11.41M &18.11G\\

		Restormer (CVPR 22)\cite{28}&23.701 &0.907 &20.797 &0.828 &\textcolor{blue}{2.688} &0.572 &2.981 &0.542 &2.485 &0.562 &26.10M &140.99G \\

		Convformer(TETCI 24)\cite{33} &23.134 &0.904 &20.894 &0.831 &2.684 &0.572 &2.946 &\textcolor{blue}{0.555} &2.315 &0.558 &25.9M &36.9G \\

		X-CAUNET(ICASSP 24)\cite{5} &22.301 &0.908 &20.675 &\textcolor{blue}{0.839}
 &2.683 &0.564 &2.922 &0.541 &2.458 &\textcolor{blue}{0.561} &31.78M &261.48G \\
\hline

		WaterMamba &\textcolor{red}{24.715} &\textcolor{red}{0.931} &\textcolor{red}{21.992} &\textcolor{red}{0.843} &\textcolor{red}{2.835} &\textcolor{red}{0.582} &\textcolor{blue}{3.057} &\textcolor{red}{0.555} &\textcolor{red}{2.767} &\textcolor{red}{0.565} &3.69M &7.53G \\
		\hline
		\specialrule{.1em}{0em}{0em} 
		\end{tabular}
	}
\end{table}
As shown in Table ~\ref{tab1}, the proposed WaterMamba method was quantitatively compared with 10 SOTA methods. WaterMamba outperformed other SOTA methods on all metrics evaluated. On the R90 dataset, compared to Semi-UIR (CVPR '23), the PSNR and SSIM of WaterMamba were improved by 0.496 and 0.029, respectively. Additionally, as depicted in Fig.~\ref{fig1} and Table ~\ref{tab1}, the number of parameters and FLOPs of WaterMamba were reduced by 22.4 and 133.5 compared to the Transformer-based Restormer (CVPR '22). This can be primarily attributed to the linear complexity state space mechanism employed in WaterMamba, which significantly reduces computational costs and makes it more suitable for real-time underwater image processing applications.

\section{Conclusion}
In this paper, we introduced a spatial-channel aware UNet architecture and proposed a novel underwater image enhancement (UIE) method, WaterMamba, which leverages the linear complexity and high-frequency modeling capability of the full-directional selective state-space model. This allows our proposed spatial-channel omnidirectional selective scan (SCOSS) block to effectively model and utilize image features at different scales. The SCOSS module consists of the spatial-channel coordinate omnidirectional selective scan (SCCOSS) block and multi-scale feedforward network (MSFFN) module. The SCOSS block utilizes the remote modeling capability of Mamba to comprehensively and effectively model image features. It demonstrates the importance of spatial and channel context in feature reconstruction, and the MSFFN further maps and regulates the flow of image information, improving the accuracy and efficiency of the network. Extensive experiments demonstrate that WaterMamba achieves good performance compared to existing methods, confirming the efficiency and effectiveness of the WaterMamba.


\appendix

\section{Appendix / supplemental material}
\subsection{Ablation Study}
 \begin{table*}[t]   
 	\renewcommand\arraystretch{1.3}
	\centering
	\fontsize{9}{8.2}\selectfont
	\caption{The quantitative results of the way of the guidance on the testing datasets were measured by the average PSNR and SSIM values.  The highest score is marked in bold red.}
	\label{tab2}
	\begin{tabular}{c|c|c|ccc|c}\hline
		\multicolumn{2}{c|}{Ablation} & \multicolumn{1}{c|}{\textbf{Baselines}} &\multicolumn{3}{c}{\textbf{Module}}& \multicolumn{1}{|c}{\textbf{Overall}} \\
		\hline
		\multicolumn{2}{c|}{\textbf{Network}}& UNet with Resblocks\cite{24}& w/o SOSS & w/o CCOSS  & w/o MSFFN & Overall\\
		\hline
		\multirow{2}{*}{R90}&PSNR&18.102 &  24.167&24.312 &24.512 &\textcolor{red}{24.715}\\
									&SSIM&0.822 &  0.918&0.922 &0.926 &\textcolor{red}{0.931}\\
		\hline
		\multicolumn{2}{c|}{\textbf{Params(M)}}& 3.35 &3.45& 3.68 & 3.56 & 3.53\\
		\hline
		\multicolumn{2}{c|}{\textbf{FLOPs(G)}}& 17.42 &7.48& 7.52 & 7.46 & 7.53\\
		\hline
		\specialrule{.1em}{0em}{0em} 
	\end{tabular}
	\label{label3}
\end{table*}

\begin{figure*}[!t]    
	\centering
	\subfigure{
		\begin{minipage}[b]{0.124\linewidth}
			\centerline{\includegraphics[height=1.3cm,width=1.9cm]{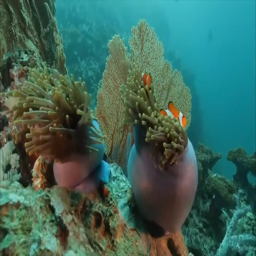}}\vspace{0.01cm}
			\centerline{\includegraphics[height=1.3cm,width=1.9cm]{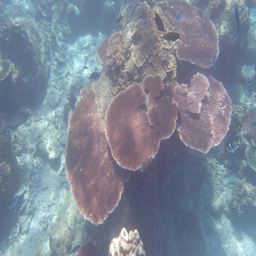}}\vspace{0.01cm} 

			\centerline{(a)}
		\end{minipage}
	}
	\subfigure{
		\begin{minipage}[b]{0.124\linewidth}
			\centerline{\includegraphics[height=1.3cm,width=1.9cm]{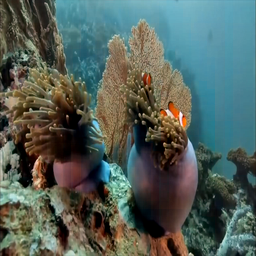}}\vspace{0.01cm}
			\centerline{\includegraphics[height=1.3cm,width=1.9cm]{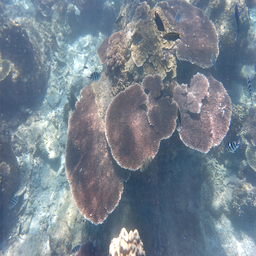}}\vspace{0.01cm} 

			\centerline{(b)}
		\end{minipage}
	}
	\subfigure{
		\begin{minipage}[b]{0.124\linewidth}
			\centerline{\includegraphics[height=1.3cm,width=1.9cm]{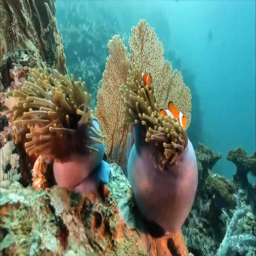}}\vspace{0.01cm}
			\centerline{\includegraphics[height=1.3cm,width=1.9cm]{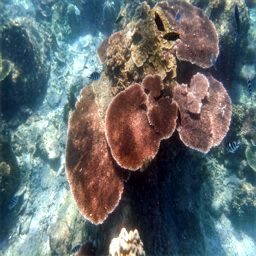}}\vspace{0.01cm} 
          
			\centerline{(c)}
		\end{minipage}
	}
	\subfigure{
		\begin{minipage}[b]{0.124\linewidth}
			\centerline{\includegraphics[height=1.3cm,width=1.9cm]{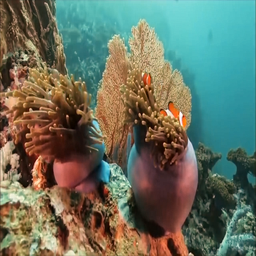}}\vspace{0.01cm}
			\centerline{\includegraphics[height=1.3cm,width=1.9cm]{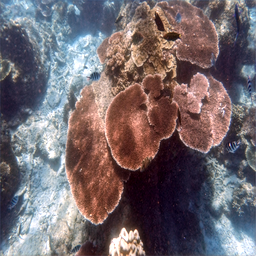}}\vspace{0.01cm}

			\centerline{(d)}
		\end{minipage}
	}
	\subfigure{
		\begin{minipage}[b]{0.124\linewidth}
			\centerline{\includegraphics[height=1.3cm,width=1.9cm]{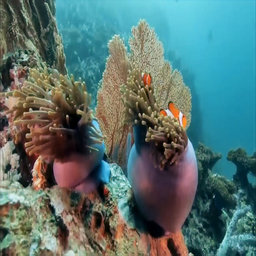}}\vspace{0.01cm}
			\centerline{\includegraphics[height=1.3cm,width=1.9cm]{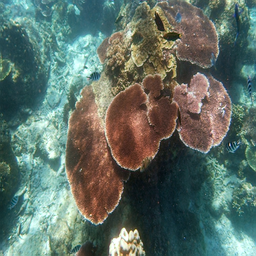}}\vspace{0.01cm}

			\centerline{(e)}
		\end{minipage}
	}
	\subfigure{
		\begin{minipage}[b]{0.124\linewidth}
			\centerline{\includegraphics[height=1.3cm,width=1.9cm]{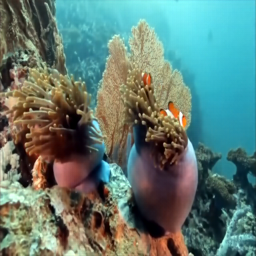}}\vspace{0.01cm}
			\centerline{\includegraphics[height=1.3cm,width=1.9cm]{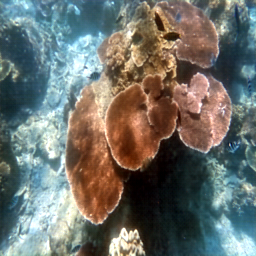}}\vspace{0.01cm}

			\centerline{(f)}
		\end{minipage}
	}
	\subfigure{
		\begin{minipage}[b]{0.124\linewidth}
			\centerline{\includegraphics[height=1.3cm,width=1.9cm]{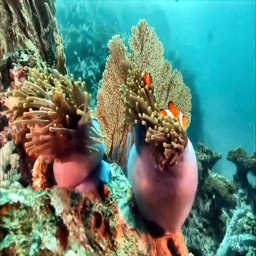}}\vspace{0.01cm}
			\centerline{\includegraphics[height=1.3cm,width=1.9cm]{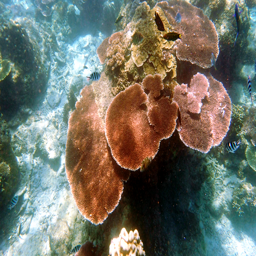}}\vspace{0.01cm}    
 
			\centerline{(g)}
		\end{minipage}
	}

\caption{ Ablation experiment on baseline, and network core component modules of the proposed WaterMamba.  (a) Raw;  (b) UNet with Resblocks\cite{24};  (c) SOSS;  (d) CCOSS  (e) MSFFN; (f) WaterMamba;  (g) reference. }
  \label{fig5}
\end{figure*}

To further analyze the role of key modules in WaterMamba, an ablation study was conducted on the proposed Mamba-based method for underwater image enhancement. As depicted in Fig. ~\ref{fig5} and Table ~\ref{tab2}, the ablation experiments systematically investigated the contribution of individual components to the overall performance. The SCOSS block and MSFFN were each removed independently from the full WaterMamba model, and the resultant impact on metrics such as PSNR and SSIM was evaluated. These diagnostic experiments provided valuable insights into the significance of multi-scale feature modeling and spatially-aware channel attention within the proposed framework. The quantitative results validate the importance of these novel components for achieving leading enhancement capability on underwater images.

\textbf{Impact of SOSS block:} To evaluate the contribution of the SOSS block, it was replaced with a standard convolutional module while keeping other parts of the network unchanged. As shown in Table ~\ref{tab2}, utilizing the SOSS block led to improvements of 6.01 dB in PSNR and 0.092 in SSIM. This validated the ability of SOSS to more effectively capture pixel-channel correlations in underwater images, as evidenced by the enhanced enhancement performance. By integrating spatial attention into channel modeling, the SOSS block facilitates multi-directional information flow between pixels and channels. This allows the network to better distinguish object features obscured by scattering and other water-specific distortions. Table ~\ref{tab2} quantitative results demonstrate the importance of the SOSS design for improving the model's feature representation capacity when handling degraded underwater images.

\textbf{Impact of CCOSS module:} To evaluate the contribution of the CCOSS block, it was replaced with a standard average pooling attention module while keeping other parts of the network unchanged. By removing the bidirectional channel scanning operation, the computational complexity of the network remained largely unchanged. However, the accuracy decreased by approximately 6.14 dB, demonstrating that the designed channel scanning mechanism significantly improves model performance with minimal additional computation. Compared to average pooling which aggregates channel-wise features, the selective channel scanning in CCOSS allows the model to learn more discriminative channel representations tailored to the challenges of underwater domains.

\textbf{Impact of MSFFN:} To evaluate the contribution of the MSFFN, it was replaced with a gated feedforward neural network module and single-scale feedforward neural network while keeping other parts of the network unchanged. As shown in Table ~\ref{tab2}, the MSFFN achieved superior performance metrics. This validation demonstrates the importance of exploiting multi-scale features for modeling the diverse characteristics of underwater images. By integrating information from multiple receptive fields, the MSFFN is better able to capture the scale-variant patterns degraded by different water conditions. These results substantiate the proposed network configuration of incorporating multi-scale analysis within the MSFFN design.

In conclusion, the above ablation experiment results validate the effectiveness of the key modules in WaterMamba, providing important evidence for further improving UIE methods. WaterMamba can generate high-quality underwater enhanced images while maintaining low complexity, striking a balance between performance and efficiency.

\subsection{Cross Datasets Studay}

To evaluate the effectiveness and generalization ability of the WaterMamba method, we conducted empirical analysis on cross datasets. Specifically, we used the following three real underwater image datasets:

1. LSUI \cite{32} Dataset: We randomly sampled 4500 pairs of real underwater images from the LSUI dataset, referred to as Train-L. The remaining 504 pairs were used as the test set, referred to as Test-L.

2. UFO-120 \cite{11} Dataset: We used 1500 pairs of real underwater images from the UFO-120 dataset, referred to as Train-UF. The remaining 120 pairs of images were used as the test set, referred to as Test-UF.

3. EUVP \cite{12} Dataset: We used 11435 pairs of real underwater images from the EUVP dataset, referred to as Train-E. 515 pairs of images were used as the test set, referred to as Test-E.

By performing cross-validation on multiple heterogeneous datasets, we comprehensively evaluated the effectiveness and generalization ability of the proposed model under different data distributions, further validating the robustness of the model.
\begin{table}
  \caption{The three cross-datasets evaluations in terms of average PSNR and SSIM Values.}

	\label{tab3}
  \centering
	\fontsize{13}{12.8}\selectfont

  \begin{tabular}{ccccccc}
    \toprule
\multirow{2}{*}{\textbf{Training}}
     &{\textbf{Test-L}}   & &{\textbf{Test-UF}}  & &{\textbf{Test-E}}          \\
    \cmidrule(r){2-7}
         & PSNR & SSIM    & PSNR & SSIM & PSNR & SSIM \\
    \midrule

\textbf{Train-L}&27.368 &0.877 &28.032 &0.856 &28.332 &0.879 \\

    \textbf{Train-UF}     &21.231 &0.802 &27.800 &0.857 &27.155 &0.866   \\
    \textbf{Train-E}   &21.998 &0.806 &27.922 &0.851 &26.925 &0.870 \\
    \bottomrule
  \end{tabular}
\end{table}

\begin{figure*}[!t]    
	\centering
	\subfigure{
		\begin{minipage}[b]{0.058\linewidth}
			\centerline{\includegraphics[width=6.3cm,height=2.1cm]{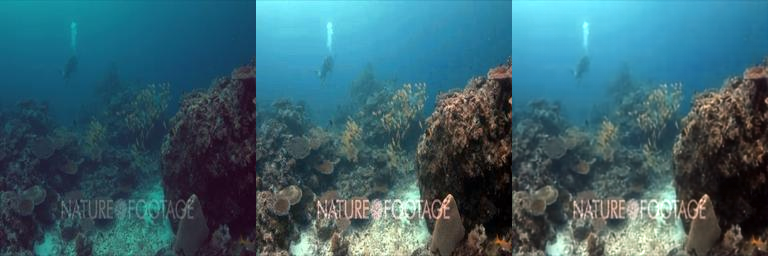} \includegraphics[width=6.3cm,height=2.1cm]{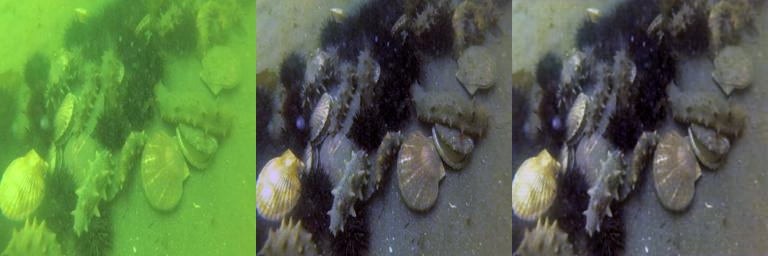}}\vspace{0.01cm}
			\centerline{\includegraphics[width=6.3cm,height=2.1cm]{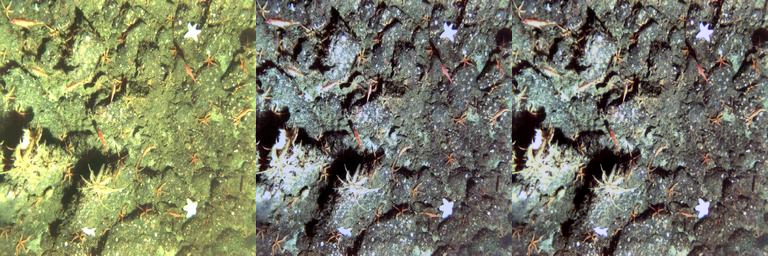} \includegraphics[width=6.3cm,height=2.1cm]{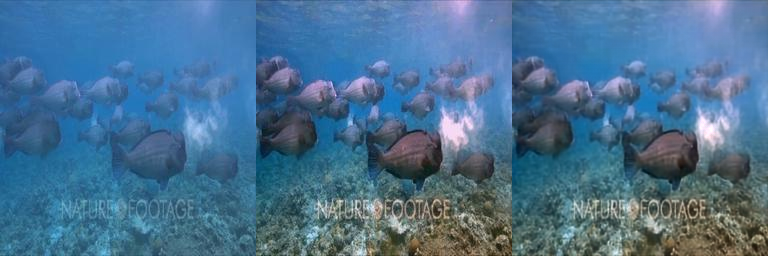}}\vspace{0.01cm}
			\centerline{LSUI}
			\centerline{\includegraphics[width=6.3cm,height=2.1cm]{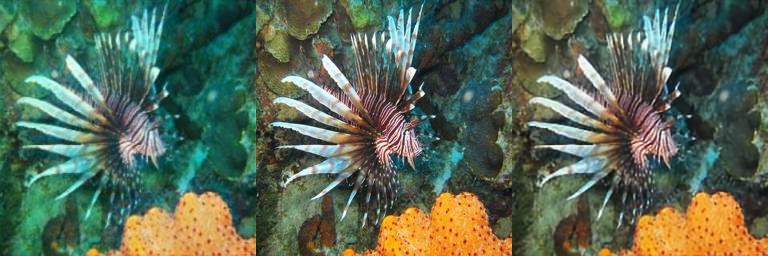} \includegraphics[width=6.3cm,height=2.1cm]{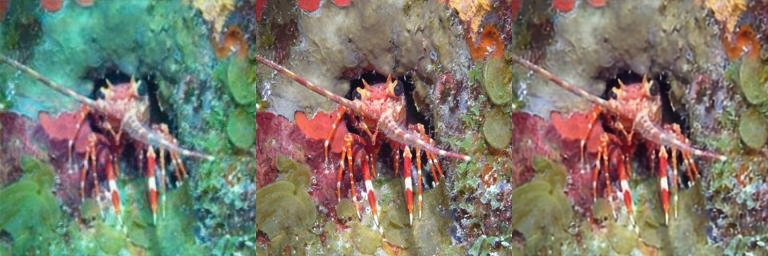}}\vspace{0.01cm}
			\centerline{\includegraphics[width=6.3cm,height=2.1cm]{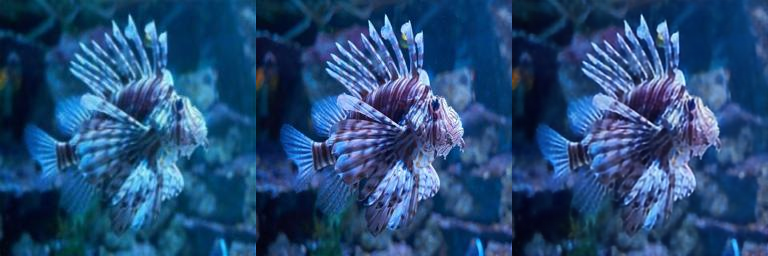} \includegraphics[width=6.3cm,height=2.1cm]{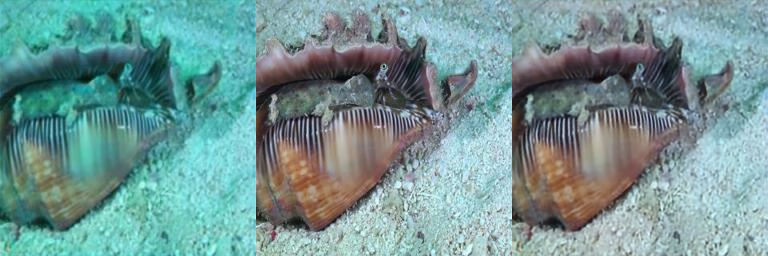}}\vspace{0.01cm}
			\centerline{UFO-120}
			\centerline{\includegraphics[width=6.3cm,height=2.1cm]{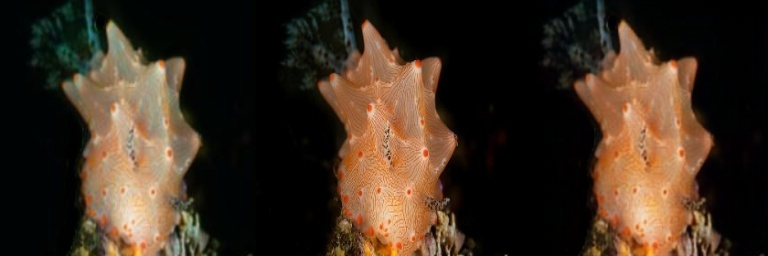} \includegraphics[width=6.3cm,height=2.1cm]{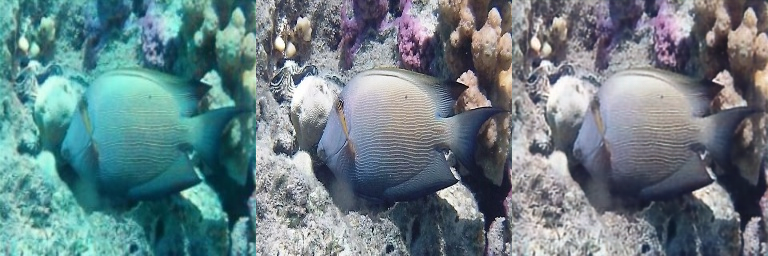}}\vspace{0.01cm}
			\centerline{\includegraphics[width=6.3cm,height=2.1cm]{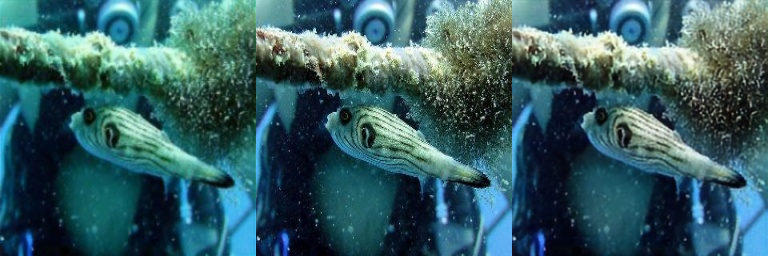} \includegraphics[width=6.3cm,height=2.1cm]{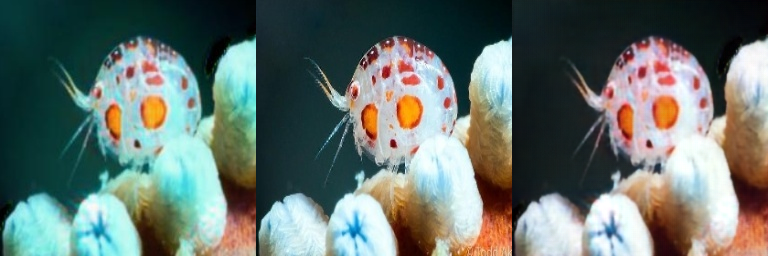}}\vspace{0.01cm}
			\centerline{EUVP}
		\end{minipage}
	}

  \caption{The restored testing results in the corresponding training dataset. In each case, from left to right is the Raw
, reference image, the result of WaterMamba.
}
  \label{fig6}
\end{figure*}
\begin{figure*}[!t]    
	\centering
	\subfigure{
		\begin{minipage}[b]{0.058\linewidth}
			\centerline{\includegraphics[width=6.3cm,height=2.1cm]{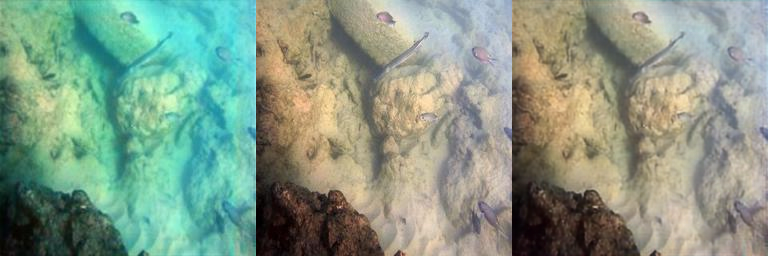} \includegraphics[width=6.3cm,height=2.1cm]{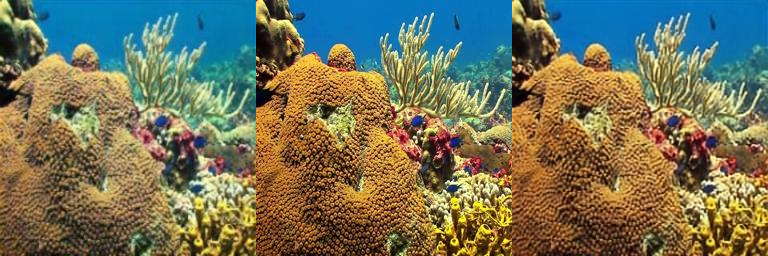}}\vspace{0.01cm}
			\centerline{Train-L, Test-UF}
			\centerline{\includegraphics[width=6.3cm,height=2.1cm]{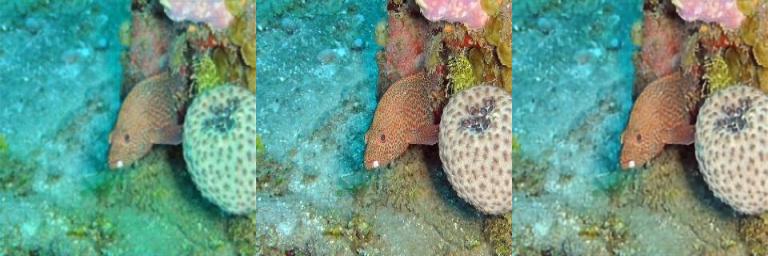} \includegraphics[width=6.3cm,height=2.1cm]{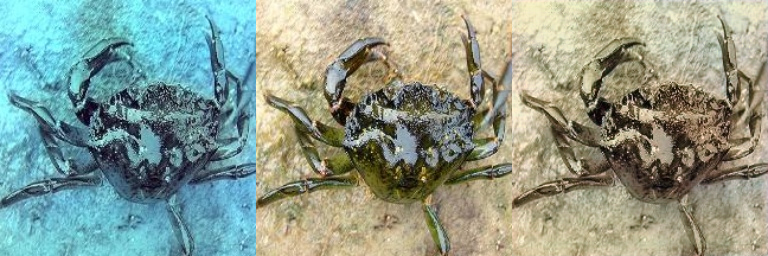}}\vspace{0.01cm}
			\centerline{Train-L, Test-E}
			\centerline{\includegraphics[width=6.3cm,height=2.1cm]{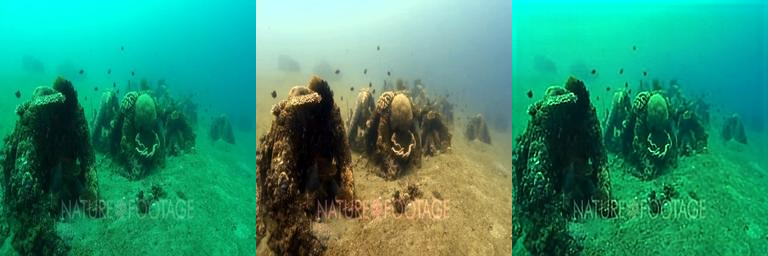} \includegraphics[width=6.3cm,height=2.1cm]{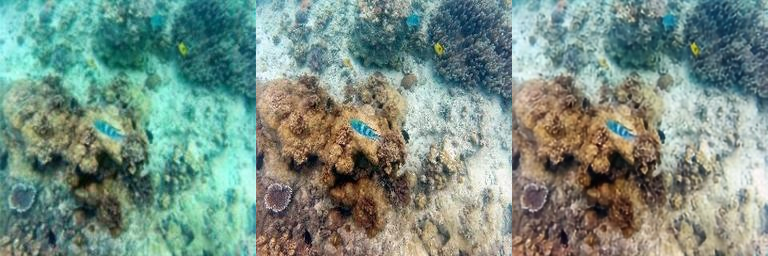}}\vspace{0.01cm}
			\centerline{Train-UF, Test-L}
			\centerline{\includegraphics[width=6.3cm,height=2.1cm]{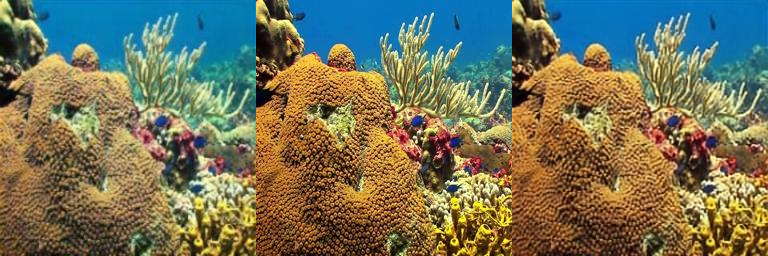} \includegraphics[width=6.3cm,height=2.1cm]{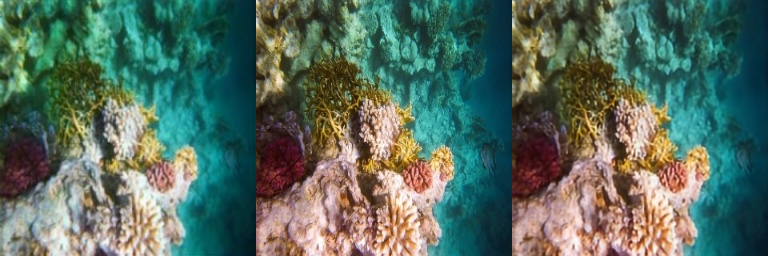}}\vspace{0.01cm}
			\centerline{Train-UF, Test-E}
			\centerline{\includegraphics[width=6.3cm,height=2.1cm]{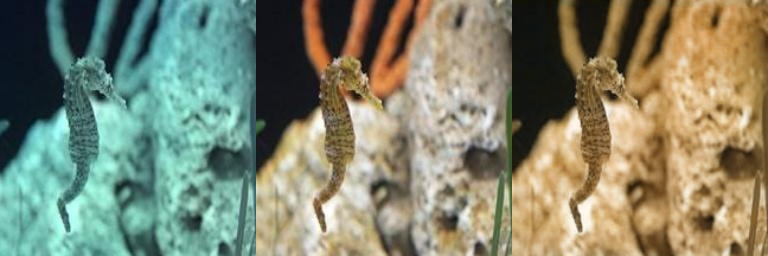} \includegraphics[width=6.3cm,height=2.1cm]{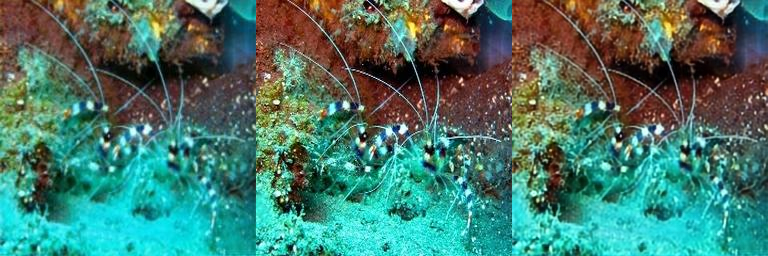}}\vspace{0.01cm}
			\centerline{Train-E, Test-L}
			\centerline{\includegraphics[width=6.3cm,height=2.1cm]{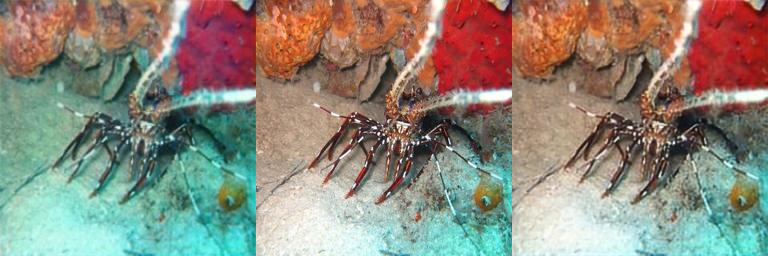} \includegraphics[width=6.3cm,height=2.1cm]{uie/wm/jc/cs/ue/2}}\vspace{0.01cm}
			\centerline{Train-E, Test-UF}
		\end{minipage}
	}

  \caption{The restored testing results in the cross-dataset. In each case, from left to right is the Raw, reference image, the result of WaterMamba.
}
  \label{fig7}
\end{figure*}

To further validate the outstanding performance of WaterMamba, four representative complex scenes were deliberately selected for comparative analysis. These scenes include the Haze scene (top left in Fig. ~\ref{fig6}), Green coverage scene (top right), Light yellow coverage scene (bottom left), and Blue coverage scene (bottom right). In addition to the image enhancement task, the model's performance on the super-resolution task was evaluated using the UFO-120 dataset. The EUVP dataset and UFO-120 dataset exhibit similarities in style, with even some overlapping data, which provides more comparable quantitative results as shown in Table ~\ref{tab3}. Challenging samples with high requirements for image texture details were specifically chosen, including the Haze scene, Green scene with interference color, Green coverage scene, and Blue coverage scene, as illustrated in Fig. ~\ref{fig7}.

In conclusion, through extensive empirical analysis on multiple heterogeneous datasets, this study fully demonstrates the outstanding performance and good generalization ability of WaterMamba in UIE and super-resolution tasks. This provides a solid foundation for the wide application of WaterMamba in underwater imaging and related fields.






\section*{NeurIPS Paper Checklist}



\begin{enumerate}

\item {\bf Claims}
    \item[] Question: Do the main claims made in the abstract and introduction accurately reflect the paper's contributions and scope?
    \item[] Answer: \answerYes{} 
    \item[] Justification: The main claims presented in the abstract and introduction of the paper accurately summarize the key contributions and scope of the research, providing a concise overview of the paper's objectives, findings, and the specific problem or topic it addresses.
    \item[] Guidelines:
    \begin{itemize}
        \item The answer NA means that the abstract and introduction do not include the claims made in the paper.
        \item The abstract and/or introduction should clearly state the claims made, including the contributions made in the paper and important assumptions and limitations. A No or NA answer to this question will not be perceived well by the reviewers. 
        \item The claims made should match theoretical and experimental results, and reflect how much the results can be expected to generalize to other settings. 
        \item It is fine to include aspirational goals as motivation as long as it is clear that these goals are not attained by the paper. 
    \end{itemize}

\item {\bf Limitations}
    \item[] Question: Does the paper discuss the limitations of the work performed by the authors?
    \item[] Answer: \answerNo{} 
    \item[] Justification: The paper does not discuss the limitations of the work performed by the authors. The paper primarily focuses on presenting the research findings, methodologies, and potential applications, without explicitly addressing any limitations or areas for improvement in the conducted work.
    \item[] Guidelines:
    \begin{itemize}
        \item The answer NA means that the paper has no limitation while the answer No means that the paper has limitations, but those are not discussed in the paper. 
        \item The authors are encouraged to create a separate "Limitations" section in their paper.
        \item The paper should point out any strong assumptions and how robust the results are to violations of these assumptions (e.g., independence assumptions, noiseless settings, model well-specification, asymptotic approximations only holding locally). The authors should reflect on how these assumptions might be violated in practice and what the implications would be.
        \item The authors should reflect on the scope of the claims made, e.g., if the approach was only tested on a few datasets or with a few runs. In general, empirical results often depend on implicit assumptions, which should be articulated.
        \item The authors should reflect on the factors that influence the performance of the approach. For example, a facial recognition algorithm may perform poorly when image resolution is low or images are taken in low lighting. Or a speech-to-text system might not be used reliably to provide closed captions for online lectures because it fails to handle technical jargon.
        \item The authors should discuss the computational efficiency of the proposed algorithms and how they scale with dataset size.
        \item If applicable, the authors should discuss possible limitations of their approach to address problems of privacy and fairness.
        \item While the authors might fear that complete honesty about limitations might be used by reviewers as grounds for rejection, a worse outcome might be that reviewers discover limitations that aren't acknowledged in the paper. The authors should use their best judgment and recognize that individual actions in favor of transparency play an important role in developing norms that preserve the integrity of the community. Reviewers will be specifically instructed to not penalize honesty concerning limitations.
    \end{itemize}

\item {\bf Theory Assumptions and Proofs}
    \item[] Question: For each theoretical result, does the paper provide the full set of assumptions and a complete (and correct) proof?
    \item[] Answer: \answerNA{} 
    \item[] Justification: The paper does not provide the full set of assumptions and a complete proof for each theoretical result. However, it compensates for this by providing experimental results and evidence to support and validate the theoretical findings, demonstrating the practical implications and effectiveness of the proposed approach or methodology.

    \item[] Guidelines:
    \begin{itemize}
        \item The answer NA means that the paper does not include theoretical results. 
        \item All the theorems, formulas, and proofs in the paper should be numbered and cross-referenced.
        \item All assumptions should be clearly stated or referenced in the statement of any theorems.
        \item The proofs can either appear in the main paper or the supplemental material, but if they appear in the supplemental material, the authors are encouraged to provide a short proof sketch to provide intuition. 
        \item Inversely, any informal proof provided in the core of the paper should be complemented by formal proofs provided in appendix or supplemental material.
        \item Theorems and Lemmas that the proof relies upon should be properly referenced. 
    \end{itemize}

    \item {\bf Experimental Result Reproducibility}
    \item[] Question: Does the paper fully disclose all the information needed to reproduce the main experimental results of the paper to the extent that it affects the main claims and/or conclusions of the paper (regardless of whether the code and data are provided or not)?
    \item[] Answer: \answerYes{} 
    \item[] Justification: The paper fully discloses all the information necessary to reproduce the main experimental results, ensuring that the main claims and conclusions of the paper can be independently verified and validated. Additionally, the authors provide supplementary materials, including code and data, to further facilitate the replication of the experiments and enhance the transparency and reproducibility of the research.
    \item[] Guidelines:
    \begin{itemize}
        \item The answer NA means that the paper does not include experiments.
        \item If the paper includes experiments, a No answer to this question will not be perceived well by the reviewers: Making the paper reproducible is important, regardless of whether the code and data are provided or not.
        \item If the contribution is a dataset and/or model, the authors should describe the steps taken to make their results reproducible or verifiable. 
        \item Depending on the contribution, reproducibility can be accomplished in various ways. For example, if the contribution is a novel architecture, describing the architecture fully might suffice, or if the contribution is a specific model and empirical evaluation, it may be necessary to either make it possible for others to replicate the model with the same dataset, or provide access to the model. In general. releasing code and data is often one good way to accomplish this, but reproducibility can also be provided via detailed instructions for how to replicate the results, access to a hosted model (e.g., in the case of a large language model), releasing of a model checkpoint, or other means that are appropriate to the research performed.
        \item While NeurIPS does not require releasing code, the conference does require all submissions to provide some reasonable avenue for reproducibility, which may depend on the nature of the contribution. For example
        \begin{enumerate}
            \item If the contribution is primarily a new algorithm, the paper should make it clear how to reproduce that algorithm.
            \item If the contribution is primarily a new model architecture, the paper should describe the architecture clearly and fully.
            \item If the contribution is a new model (e.g., a large language model), then there should either be a way to access this model for reproducing the results or a way to reproduce the model (e.g., with an open-source dataset or instructions for how to construct the dataset).
            \item We recognize that reproducibility may be tricky in some cases, in which case authors are welcome to describe the particular way they provide for reproducibility. In the case of closed-source models, it may be that access to the model is limited in some way (e.g., to registered users), but it should be possible for other researchers to have some path to reproducing or verifying the results.
        \end{enumerate}
    \end{itemize}

\item {\bf Open access to data and code}
    \item[] Question: Does the paper provide open access to the data and code, with sufficient instructions to faithfully reproduce the main experimental results, as described in supplemental material?
    \item[] Answer: \answerYes{} 
    \item[] Justification: The paper provides open access to the data and code, along with sufficient instructions, allowing for the faithful reproduction of the main experimental results as described in the supplemental material. This ensures transparency and enables other researchers to replicate and build upon the findings presented in the paper.
    \item[] Guidelines:
    \begin{itemize}
        \item The answer NA means that paper does not include experiments requiring code.
        \item Please see the NeurIPS code and data submission guidelines (\url{https://nips.cc/public/guides/CodeSubmissionPolicy}) for more details.
        \item While we encourage the release of code and data, we understand that this might not be possible, so “No” is an acceptable answer. Papers cannot be rejected simply for not including code, unless this is central to the contribution (e.g., for a new open-source benchmark).
        \item The instructions should contain the exact command and environment needed to run to reproduce the results. See the NeurIPS code and data submission guidelines (\url{https://nips.cc/public/guides/CodeSubmissionPolicy}) for more details.
        \item The authors should provide instructions on data access and preparation, including how to access the raw data, preprocessed data, intermediate data, and generated data, etc.
        \item The authors should provide scripts to reproduce all experimental results for the new proposed method and baselines. If only a subset of experiments are reproducible, they should state which ones are omitted from the script and why.
        \item At submission time, to preserve anonymity, the authors should release anonymized versions (if applicable).
        \item Providing as much information as possible in supplemental material (appended to the paper) is recommended, but including URLs to data and code is permitted.
    \end{itemize}

\item {\bf Experimental Setting/Details}
    \item[] Question: Does the paper specify all the training and test details (e.g., data splits, hyperparameters, how they were chosen, type of optimizer, etc.) necessary to understand the results?
    \item[] Answer: \answerYes{} 
    \item[] Justification: the paper specifies all the training and test details necessary to understand the results, including information about data splits, hyperparameters, the process of selecting hyperparameters, the type of optimizer used, and any other relevant details. This comprehensive disclosure enables readers to have a clear understanding of the experimental setup and facilitates reproducibility of the results.
    \item[] Guidelines:
    \begin{itemize}
        \item The answer NA means that the paper does not include experiments.
        \item The experimental setting should be presented in the core of the paper to a level of detail that is necessary to appreciate the results and make sense of them.
        \item The full details can be provided either with the code, in appendix, or as supplemental material.
    \end{itemize}

\item {\bf Experiment Statistical Significance}
    \item[] Question: Does the paper report error bars suitably and correctly defined or other appropriate information about the statistical significance of the experiments?
    \item[] Answer: \answerNA{} 
    \item[] Justification: there is no reporting of error bars or statistical significance information.
    \item[] Guidelines:
    \begin{itemize}
        \item The answer NA means that the paper does not include experiments.
        \item The authors should answer "Yes" if the results are accompanied by error bars, confidence intervals, or statistical significance tests, at least for the experiments that support the main claims of the paper.
        \item The factors of variability that the error bars are capturing should be clearly stated (for example, train/test split, initialization, random drawing of some parameter, or overall run with given experimental conditions).
        \item The method for calculating the error bars should be explained (closed form formula, call to a library function, bootstrap, etc.)
        \item The assumptions made should be given (e.g., Normally distributed errors).
        \item It should be clear whether the error bar is the standard deviation or the standard error of the mean.
        \item It is OK to report 1-sigma error bars, but one should state it. The authors should preferably report a 2-sigma error bar than state that they have a 96\% CI, if the hypothesis of Normality of errors is not verified.
        \item For asymmetric distributions, the authors should be careful not to show in tables or figures symmetric error bars that would yield results that are out of range (e.g. negative error rates).
        \item If error bars are reported in tables or plots, The authors should explain in the text how they were calculated and reference the corresponding figures or tables in the text.
    \end{itemize}

\item {\bf Experiments Compute Resources}
    \item[] Question: For each experiment, does the paper provide sufficient information on the computer resources (type of compute workers, memory, time of execution) needed to reproduce the experiments?
    \item[] Answer: \answerNo{} 
    \item[] Justification: The paper does not provide sufficient information on the computer resources, such as the type of compute workers, memory, and time of execution needed to reproduce the experiments. However, the paper compensates for this by providing details on the computational complexity of the proposed methods, giving an indication of the required computational resources without specifying the exact hardware or execution times.
    \item[] Guidelines:
    \begin{itemize}
        \item The answer NA means that the paper does not include experiments.
        \item The paper should indicate the type of compute workers CPU or GPU, internal cluster, or cloud provider, including relevant memory and storage.
        \item The paper should provide the amount of compute required for each of the individual experimental runs as well as estimate the total compute. 
        \item The paper should disclose whether the full research project required more compute than the experiments reported in the paper (e.g., preliminary or failed experiments that didn't make it into the paper). 
    \end{itemize}
    
\item {\bf Code Of Ethics}
    \item[] Question: Does the research conducted in the paper conform, in every respect, with the NeurIPS Code of Ethics \url{https://neurips.cc/public/EthicsGuidelines}?
    \item[] Answer: \answerYes{} 
    \item[] Justification: The research conducted in the paper conforms to the NeurIPS Code of Ethics, as outlined in the provided URL. The paper adheres to the ethical practices and guidelines specified in the NeurIPS Code of Ethics during the research process.
    \item[] Guidelines:
    \begin{itemize}
        \item The answer NA means that the authors have not reviewed the NeurIPS Code of Ethics.
        \item If the authors answer No, they should explain the special circumstances that require a deviation from the Code of Ethics.
        \item The authors should make sure to preserve anonymity (e.g., if there is a special consideration due to laws or regulations in their jurisdiction).
    \end{itemize}

\item {\bf Broader Impacts}
    \item[] Question: Does the paper discuss both potential positive societal impacts and negative societal impacts of the work performed?
    \item[] Answer: \answerNo{} 
    \item[] Justification: The paper solely emphasizes the positive societal impacts of the work performed, omitting any discussion of potential negative consequences or societal drawbacks.
    \item[] Guidelines:
    \begin{itemize}
        \item The answer NA means that there is no societal impact of the work performed.
        \item If the authors answer NA or No, they should explain why their work has no societal impact or why the paper does not address societal impact.
        \item Examples of negative societal impacts include potential malicious or unintended uses (e.g., disinformation, generating fake profiles, surveillance), fairness considerations (e.g., deployment of technologies that could make decisions that unfairly impact specific groups), privacy considerations, and security considerations.
        \item The conference expects that many papers will be foundational research and not tied to particular applications, let alone deployments. However, if there is a direct path to any negative applications, the authors should point it out. For example, it is legitimate to point out that an improvement in the quality of generative models could be used to generate deepfakes for disinformation. On the other hand, it is not needed to point out that a generic algorithm for optimizing neural networks could enable people to train models that generate Deepfakes faster.
        \item The authors should consider possible harms that could arise when the technology is being used as intended and functioning correctly, harms that could arise when the technology is being used as intended but gives incorrect results, and harms following from (intentional or unintentional) misuse of the technology.
        \item If there are negative societal impacts, the authors could also discuss possible mitigation strategies (e.g., gated release of models, providing defenses in addition to attacks, mechanisms for monitoring misuse, mechanisms to monitor how a system learns from feedback over time, improving the efficiency and accessibility of ML).
    \end{itemize}
    
\item {\bf Safeguards}
    \item[] Question: Does the paper describe safeguards that have been put in place for responsible release of data or models that have a high risk for misuse (e.g., pretrained language models, image generators, or scraped datasets)?
    \item[] Answer: \answerNo{} 
    \item[] Justification: The paper lacks any mention or description of safeguards implemented for responsible release of high-risk data or models, such as pretrained language models, image generators, or scraped datasets, thus failing to address potential ethical concerns regarding their misuse.
    \item[] Guidelines:
    \begin{itemize}
        \item The answer NA means that the paper poses no such risks.
        \item Released models that have a high risk for misuse or dual-use should be released with necessary safeguards to allow for controlled use of the model, for example by requiring that users adhere to usage guidelines or restrictions to access the model or implementing safety filters. 
        \item Datasets that have been scraped from the Internet could pose safety risks. The authors should describe how they avoided releasing unsafe images.
        \item We recognize that providing effective safeguards is challenging, and many papers do not require this, but we encourage authors to take this into account and make a best faith effort.
    \end{itemize}

\item {\bf Licenses for existing assets}
    \item[] Question: Are the creators or original owners of assets (e.g., code, data, models), used in the paper, properly credited and are the license and terms of use explicitly mentioned and properly respected?
    \item[] Answer: \answerYes{} 
    \item[] Justification: The paper thoroughly acknowledges and properly credits the creators or original owners of assets, including code, data, and models, used in the research. Additionally, it explicitly mentions and respects the licenses and terms of use associated with these assets, ensuring ethical and legal compliance.
    \item[] Guidelines:
    \begin{itemize}
        \item The answer NA means that the paper does not use existing assets.
        \item The authors should cite the original paper that produced the code package or dataset.
        \item The authors should state which version of the asset is used and, if possible, include a URL.
        \item The name of the license (e.g., CC-BY 4.0) should be included for each asset.
        \item For scraped data from a particular source (e.g., website), the copyright and terms of service of that source should be provided.
        \item If assets are released, the license, copyright information, and terms of use in the package should be provided. For popular datasets, \url{paperswithcode.com/datasets} has curated licenses for some datasets. Their licensing guide can help determine the license of a dataset.
        \item For existing datasets that are re-packaged, both the original license and the license of the derived asset (if it has changed) should be provided.
        \item If this information is not available online, the authors are encouraged to reach out to the asset's creators.
    \end{itemize}

\item {\bf New Assets}
    \item[] Question: Are new assets introduced in the paper well documented and is the documentation provided alongside the assets?
    \item[] Answer: \answerNA{} 
    \item[] Justification: The paper does not introduce any new assets, hence there is no documentation provided alongside them. This may be because the research primarily builds upon existing assets rather than creating new ones.
    \item[] Guidelines:
    \begin{itemize}
        \item The answer NA means that the paper does not release new assets.
        \item Researchers should communicate the details of the dataset/code/model as part of their submissions via structured templates. This includes details about training, license, limitations, etc. 
        \item The paper should discuss whether and how consent was obtained from people whose asset is used.
        \item At submission time, remember to anonymize your assets (if applicable). You can either create an anonymized URL or include an anonymized zip file.
    \end{itemize}

\item {\bf Crowdsourcing and Research with Human Subjects}
    \item[] Question: For crowdsourcing experiments and research with human subjects, does the paper include the full text of instructions given to participants and screenshots, if applicable, as well as details about compensation (if any)? 
    \item[] Answer: \answerNo{} 
    \item[] Justification: The paper does not include crowdsourcing experiments or research with human subjects; therefore, there are no instructions provided to participants, screenshots, or details about compensation, as this aspect of research was not conducted.
    \item[] Guidelines:
    \begin{itemize}
        \item The answer NA means that the paper does not involve crowdsourcing nor research with human subjects.
        \item Including this information in the supplemental material is fine, but if the main contribution of the paper involves human subjects, then as much detail as possible should be included in the main paper. 
        \item According to the NeurIPS Code of Ethics, workers involved in data collection, curation, or other labor should be paid at least the minimum wage in the country of the data collector. 
    \end{itemize}

\item {\bf Institutional Review Board (IRB) Approvals or Equivalent for Research with Human Subjects}
    \item[] Question: Does the paper describe potential risks incurred by study participants, whether such risks were disclosed to the subjects, and whether Institutional Review Board (IRB) approvals (or an equivalent approval/review based on the requirements of your country or institution) were obtained?
    \item[] Answer: \answerNA{} 
    \item[] Justification: The paper does not describe potential risks incurred by study participants, nor does it discuss whether such risks were disclosed to subjects or whether Institutional Review Board (IRB) approvals or equivalent approvals were obtained. This is because the research does not involve human subjects or experimentation with potential risks to participants.
    \item[] Guidelines:
    \begin{itemize}
        \item The answer NA means that the paper does not involve crowdsourcing nor research with human subjects.
        \item Depending on the country in which research is conducted, IRB approval (or equivalent) may be required for any human subjects research. If you obtained IRB approval, you should clearly state this in the paper. 
        \item We recognize that the procedures for this may vary significantly between institutions and locations, and we expect authors to adhere to the NeurIPS Code of Ethics and the guidelines for their institution. 
        \item For initial submissions, do not include any information that would break anonymity (if applicable), such as the institution conducting the review.
    \end{itemize}

\end{enumerate}

\end{document}